\documentclass[10pt,journal,compsoc]{IEEEtran}

\usepackage{graphicx}
\usepackage[many]{tcolorbox}

\usepackage{mathtools,amssymb,latexsym,amsfonts,stmaryrd}
\usepackage{pbox}
\usepackage{xfrac}
\usepackage{booktabs}
\usepackage{xcolor}
\usepackage{threeparttable}
\usepackage{amsthm}
\usepackage{hyperref}
\usepackage{multirow}
\usepackage{tikz}
\usepackage{mathrsfs}
\usepackage{mathpartir}
\usepackage[ruled,linesnumbered]{algorithm2e}
\usepackage{url}
\usepackage{syntax}
\usepackage{framed}
\usepackage[noend]{algpseudocode}
\usepackage{flushend}
\usepackage{listings}
\usepackage{moresize}
\usepackage{wrapfig}
 
\usepackage{seqsplit}
\usepackage{alltt}
\usepackage{xspace}
\usepackage{enumitem}
\usepackage{pifont}% http://ctan.org/pkg/pifont

\DeclareMathAlphabet{\mathcal}{OMS}{cmsy}{m}{n}

\usepackage{amsmath, listings, amsthm, amssymb, proof, xspace}

\usepackage{thmtools}

\declaretheoremstyle[spaceabove=\topsep,notefont=\normalfont\itshape]{mystyle}

\newcommand{\revise}[2]{{\color{red}{\ifx&#1&\else- #1\fi}} {\color{ForestGreen}{\ifx&#2&\else+ #2\fi}}}%
\renewcommand{\revise}[2]{#2}%

\usetikzlibrary{arrows,patterns, decorations.pathreplacing}

\newcommand{\F}{Fig.}

\newcommand{\T}{Table}
\renewcommand{\S}{Sec.}
\newcommand{\A}{Alg.}

\newcommand{\mr}[1]{{#1}}
\newcommand{\ignore}[1]{}

\newcommand{\ts}{$\mathcal{S}$\xspace}

\newtheorem{assumption}{Assumption}
\newtheorem{theorem}{Theorem}

\newtheorem{lemma}{Lemma}
\newtheorem{definition}{Definition}
\newtheorem{subdefinition}{Definition}[definition]

\newcommand{\parh}[1]{\noindent\textbf{#1}}
\newcommand{\Asmp}{Assumption}

\newcommand{\Lem}{Lemma}
\newcommand{\Thm}{Theorem}

\newcommand{\Df}{Def.}

\lstdefinestyle{base}{
  moredelim=**[is][\color{red}]{@}{@},
  escapeinside={<@}{@>}
}

\newcommand{\tool}{\textsc{DeepWalk}}
\newcommand{\divv}{\textsc{Div}}
\newcommand{\vall}{\textsc{Val}}
\newcommand{\gen}{\textsc{DataGen}}

\usepackage{amssymb}% http://ctan.org/pkg/amssymb
\usepackage{pifont}% http://ctan.org/pkg/pifont

\newcommand\DejaVuttfamily{%
  \fontfamily{DejaVuSansMono-TLF}\selectfont }

\lstdefinestyle{base}{
  moredelim=**[is][\color{red}]{@}{@},
  escapeinside={<@}{@>}
}

\lstdefinelanguage
   [x64]{Assembler}     % add a "x64" dialect of Assembler
   [x86masm]{Assembler} % based on the "x86masm" dialect
   {morekeywords={CDQE,CQO,CMPSQ,CMPXCHG16B,JRCXZ,LODSQ,MOVSXD, %
                  POPFQ,PUSHFQ,SCASQ,STOSQ,IRETQ,RDTSCP,SWAPGS, %
                  rax,rdx,rcx,rbx,rsi,rdi,rsp,rbp, %
                  r8,r8d,r8w,r8b,r9,r9d,r9w,r9b}} % etc.

\usepackage[compact]{titlesec}

\usepackage{listings}
\usepackage{fancyvrb}
\usepackage{color}
\definecolor{lightgray}{rgb}{.9,.9,.9}
\definecolor{darkgray}{rgb}{.4,.4,.4}
\definecolor{purple}{rgb}{0.65, 0.12, 0.82}
\definecolor{commentgreen}{RGB}{63,127,95}

\colorlet{myPurple}{blue!40!red}
\definecolor{myOrange}{RGB}{255,192,0}

\newcommand{\enc}[1]{$\phi^{*}_{\theta}$}
\newcommand{\dec}[1]{$\psi^{*}_{\theta}$}

\lstdefinelanguage{Solidity}{
  keywords={len,delete,int,void,payable, public, event, contract, typeof, new, true, false, catch, function, return, null, catch, switch, var, if, in, while, do, else, case, break,struct,const,socklen_t,sa_familty_t,char,sockaddr},
  keywordstyle=\color{violet}\bfseries,
  ndkeywords={class, export, boolean, throw, implements, import, this},
  ndkeywordstyle=\color{darkgray}\bfseries,
  identifierstyle=\color{black},
  sensitive=false,
  comment=[l]{//},
  escapeinside={(*@}{@*)},          % if you want to add LaTeX within your code
  morecomment=[s]{/*}{*/},
  commentstyle=\color{commentgreen}\ttfamily,
  stringstyle=\color{red}\ttfamily,
  morestring=[b]',
  morestring=[b]"
}

\newcommand{\rnum}[1]{\uppercase\expandafter{\romannumeral #1\relax}}

\usepackage{xcolor}

\algnewcommand{\LeftComment}[1]{\Statex \(\triangleright\) #1}

\definecolor{pptbrown}{RGB}{132,60,12}
\definecolor{pptgreen}{RGB}{169,209,142}
\definecolor{pptyellow}{RGB}{255,192,0}

\let\OLDthebibliography\thebibliography
\renewcommand\thebibliography[1]{
  \OLDthebibliography{#1}
  \setlength{\parskip}{0pt}
  \setlength{\itemsep}{0pt plus 0.1ex}
}

\definecolor{pptred}{RGB}{176,35,24}
\definecolor{pptblue}{RGB}{194,214,236}
\definecolor{pptblue1}{RGB}{31,78,121}

\definecolor{pptgreen1}{RGB}{78,173,91}
\definecolor{pptred1}{RGB}{192,0,0}

\definecolor{pptyellow1}{RGB}{203,195,167}
\definecolor{pptgreen2}{RGB}{184,192,176}

\begin{document}

\title{Provably Valid and Diverse Mutations of Real-World
Media Data for DNN Testing\thanks{Preprint under review.}}

\author{
  {\rm Yuanyuan Yuan, Qi Pang, Shuai Wang~\IEEEmembership{Member,~IEEE,}}\\
  The Hong Kong University of Science and Technology\\
  \textit{\{yyuanaq,  qpangaa, shuaiw\}@cse.ust.hk}
}

\IEEEtitleabstractindextext{
\begin{abstract}

  Deep neural networks (DNNs) often accept high-dimensional media data (e.g.,
  photos, text, and audio) and understand their perceptual content (e.g., a
  cat). To test DNNs, \textit{diverse} inputs are needed to trigger
  mis-predictions. Some preliminary works use byte-level mutations or
  domain-specific filters (e.g., foggy), whose enabled mutations may be limited
  and likely error-prone. State-of-the-art (SOTA) works employ deep generative
  models to generate (infinite) inputs. Also, to keep the mutated inputs
  perceptually \textit{valid} (e.g., a cat remains a ``cat'' after mutation),
  existing efforts rely on imprecise and less generalizable heuristics.
  
  This study revisits two key objectives in media input mutation --- perception
  diversity (\divv) and validity (\vall) --- in a rigorous manner based on manifold,
  a well-developed theory capturing perceptions of high-dimensional media data in a low-dimensional
  space. We show important results that \divv\ and \vall\ inextricably bound
  each other, and prove that SOTA generative model-based methods
  fundamentally fail to mutate \textit{real-world media data} (either
  sacrificing \divv\ or \vall). In contrast, we discuss the feasibility of mutating
  real-world media data with provably high \divv\ and \vall\ based on manifold.

  Following, we concretize the technical solution of mutating media data of
  various formats (images, audios, text) via a \textit{unified} manner based on
  manifold. Specifically, when media data are projected into a low-dimensional
  manifold, the data can be mutated by walking on the manifold with certain
  directions and step sizes. When contrasted with the input data, the mutated
  data exhibit encouraging \divv\ in the perceptual traits (e.g., lying vs. standing dog)
  while retaining reasonably high \vall\ (i.e., a dog remains a dog). 
  
  We implement our techniques in \tool\ for testing DNNs. \tool\ constructs
  manifolds for media data offline. In online testing, \tool\ walks on manifolds
  to generate mutated media data with provably high \divv\ and \vall. 
  Our evaluation tests DNNs executing various tasks (e.g., classification,
  self-driving, machine translation) and media data of different types (image,
  audio, text). \tool\ outperforms prior methods in terms of the testing comprehensiveness
  and can find more error-triggering inputs with higher quality. The
  tested DNNs, after repaired using \tool's findings, exhibit better accuracy. 

\end{abstract}

\begin{IEEEkeywords}
  Deep learning testing, deep neural networks, metamorphic testing
\end{IEEEkeywords}}

\maketitle

\IEEEdisplaynontitleabstractindextext
\IEEEpeerreviewmaketitle

\section{Introduction}
\label{sec:introduciton}

\IEEEPARstart{D}{eep} neural networks (DNNs) primarily process high-dimensional
media data, such as image/text/audio, with perceptually meaningful contents
(e.g., a dog) that are human-perceivable. It is widely accepted that DNNs make
decisions based on perceptual contents in media inputs. Despite their rapid
advancement and use in safety-critical scenarios, DNNs, like traditional software,
can make incorrect predictions with fatal consequences. 

To test DNNs, we often need to launch massive media input mutations to trigger
DNN prediction errors. In this regard, it imposes a key demand on the perception
\textit{diversity} and perception \textit{validity} of mutated media inputs.
Media inputs are desirable to be diverse, thus better stressing the DNN and
exposing its faults. Moreover, media inputs are well-formed (e.g., RGB image
pixels range from $[0,255]$) and contain perception-level constraints. DNN
faults triggered by ill-formed inputs or inputs with broken perceptions are
less useful, as they are not specifically considered by DNNs. Note that DNNs do
not feature ``input check,'' and they generally assume that test inputs are
correlated with training inputs, and both test and training inputs are 
real-world meaningful media data.

\begin{figure*}[!ht]
  \centering
  \includegraphics[width=1\linewidth]{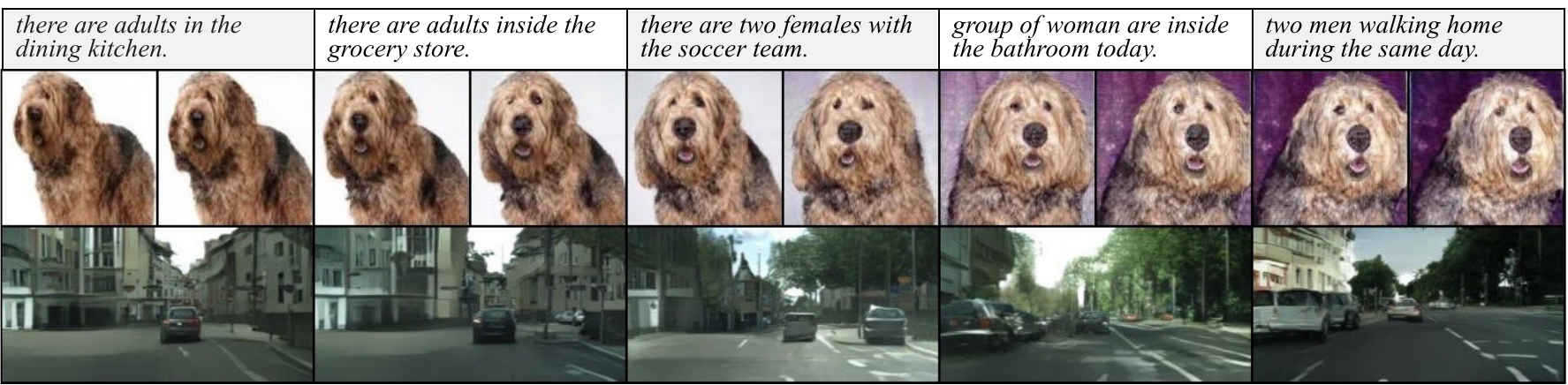}
  %\vspace*{-5pt}
  \caption{Perceptual-level mutations of \tool\ by walking towards one direction
  with a fixed step size on manifolds. \tool\ can mutate \textit{images},
  \textit{audio}, and \textit{natural language text}; see~\cite{snapshot} for
  more cases.}
%  \vspace*{-10pt}
  \label{fig:demo}
\end{figure*}

Existing research has tested DNNs by randomly changing their input data bytes
(e.g., adding random noise to images)~\cite{pei2017deepxplore}. These methods
mutate only limited bytes per input, hardly affecting the visual diversity or
may generate inputs with broken content. Recent works use weather filters or
knowledge transfer techniques~\cite{tian2018deeptest,zhang2018deeproad} to
mutate media data. However, they are limited to specific scenarios or data types
like traffic scene images~\cite{odena2018tensorfuzz, wang2019adversarial,
dwarakanath2018identifying, nakajima2019generating} or natural language
text~\cite{galhotra2017fairness, udeshi2018automated,
ma2020metamorphic,he2020structure}. 
Even worse, to generate perceptually valid image/text, existing works rely on
heuristics or domain-specific checks~\cite{xie2018coverage,zhang2018deeproad},
such as bounding the maximal number of mutated pixels. These validation methods
are often less generalizable and heavily rely on pre-trained oracle models.
Worse, they may overlook format-invalid data; as shown in
\S~\ref{sec:evaluation}.

The state-of-the-art (SOTA) works,~\cite{kang2020sinvad} and
\cite{dola2021distribution}
characterize all images via a unified latent space, and they generate media data
using advanced deep generative models. With generative models, they can generate
unlimited and diverse media data as the test inputs of DNNs. 
However, this paper demonstrates, via both rigorous proof and empirical
evaluation, that SOTA works are \textit{only} applicable
to toy datasets like MNIST~\cite{deng2012mnist} (i.e., images of handwritten digits).

This work aims to enhance media data mutation, a key step in DNN testing.
Despite having been widely discussed in existing literatures, this task is yet
rigorously formulated, leading to less practical or spurious methods. Therefore,
we focus on the two key objectives: \textit{perception diversity (\divv)} and
\textit{perception validity (\vall)}. We formulate \divv\ and \vall\ quantitatively based on manifold, a
well-established concept in representation learning, that encapsulates perceptions of high-dimensional media data
in low-dimensional spaces~\cite{bengio2013representation,zhu2018image}.
Then, we discover important and inextricable constraints between \divv\ and
\vall: \vall\ is bounded by \divv, and when images in different manifolds are
mapped to one latent space (as in SOTA works~\cite{kang2020sinvad,dola2021distribution}),
the bound is \textit{provably negligible}, greatly impeding mutating media data while
retaining a reasonable \vall. This way, we respectfully rebut SOTA works by proving that
they inevitably sacrifice \vall\ when mutating \textit{real-world media data}. In
contrast, we show the feasibility of mutating real-world media data with provably
high \divv\ and \vall\ on the basis of manifold. 

Concretely, we propose the technical approach of mutating media data by
``walking'' on their manifolds, enabling format-agnostic mutations toward
image/audio/text without domain knowledge. As in \F~\ref{fig:demo}, media
data can be mutated by first projecting them into respective manifolds, and then
moving them along the manifold in specific directions and step sizes. Contrasted
with the original data, the mutated data exhibit diverse changes in the
perceptual traits (e.g., left-oriented vs. right-oriented face). It also retains
the perceptual-level validity (e.g., a dog is still a dog). 

We implement \tool, an efficient and practical DNN testing framework. \tool\
constructs a set of manifolds $\mathbb{M}$ for media data in an \textit{offline}
phase. Then, \tool\ launches an \textit{online} testing phase, where it walks on
each $\mathcal{M} \subset \mathbb{M}$ in various directions and step sizes,
\mr{under the feedback of standard DNN testing objectives such as
structural coverage~\cite{pei2017deepxplore}, distribution-wise coverage~\cite{yuan2023revisiting},
cluster-based coverage~\cite{odena2018tensorfuzz}, or black-box objective.}
The coordinates covered on the manifold are
mapped back to high-dimensional space. Thus, media data with diverse but
nevertheless valid perceptions can be used to test DNNs and reveal their
prediction inconsistencies.

\begin{figure*}[!ht]
    \centering
    % \vspace*{-5pt}
    \includegraphics[width=1.0\linewidth]{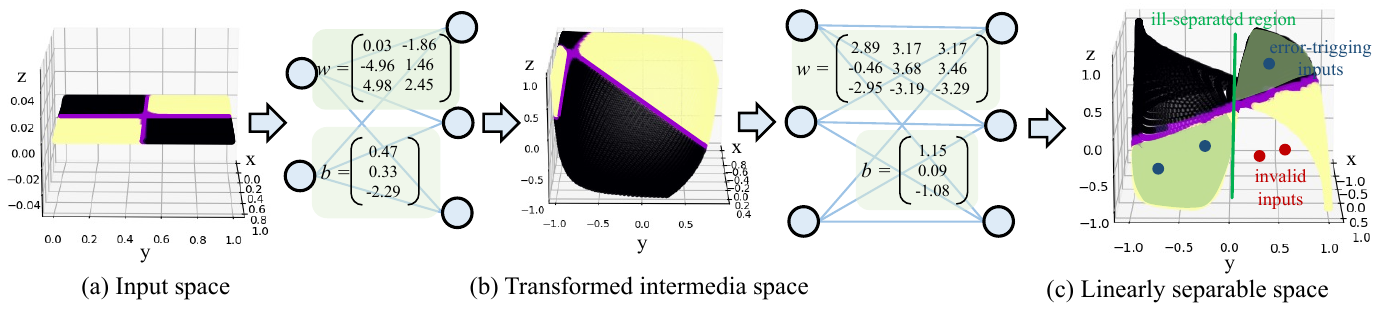}
    %\vspace*{-10pt}
    \caption{A two-layer DNN that solves the XOR problem and predicts outputs
    for $(x > 0.5) \oplus (y > 0.5)$. The input space from (a)
    is progressively transformed as linearly separable in (c). In (b) and (c),
    each dimension (i.e., $x$, $y$ and $z$) denotes one neuron and its outputs.}
    \vspace*{-5pt}
    \label{fig:func}
\end{figure*}

\mr{Nineteen popular DNNs that analyze media data in image, audio, and text formats are
evaluated. }The working contexts range from generic (e.g., image classification)
to specific (e.g., autonomous driving). We compared \tool\ with previous generative
model-based methods (e.g., DeepRoad~\cite{zhang2018deeproad}), and four de facto feedback-driven
testing frameworks --- DeepHunter~\cite{xie2018coverage}, TensorFuzz~\cite{odena2018tensorfuzz},
DeepSmart~\cite{demir2019deepsmartfuzzer}, and DeepTest~\cite{tian2018deeptest}.
\mr{\tool\ has considerably better performance than the previous works in 92 out of 95 quantitative evaluations.}
In comparison with all previous works, \tool\ also demonstrates promising results by
recognizing more and diverse error-triggering inputs. Media data mutated by \tool\ are
of better quality and can be used to better repair DNNs, outperforming previous
works. By mutating image, text, and audio inputs, we demonstrate the versatility
of \tool\ in testing DNNs for image classification, autonomous driving,
machine translation, and audio classification. Ablation studies also prove the
superiority of \tool's design considerations. In sum, this work makes the following
major contributions:

\begin{itemize}[leftmargin=6mm,noitemsep,topsep=0pt]
\item We rigorously formulate two key objectives of input mutations in DNN testing,
perception diversity (\divv) and validity (\vall), based on manifold. We prove important
results that \divv\ and \vall\ inextricably bound each other, and respectfully rebut
SOTA works by proving their incapability in front of real-world media data.

\item Following, we propose practically feasible solutions to mutate media data
  with provably high \divv\ and \vall\ based on manifold. We concretize this
  design by first recasting media data into manifolds and then performing
  perceptual-level mutations. 

\item We implement \tool\ to allow mutating media data of several formats
  without any domain knowledge or templates. \tool\ incorporates design
  principles and optimizations to explore perception changes that can trigger
  DNN faults under feedback of off-the-shelf DNN testing objectives.

\item We evaluate \tool\ using real-world DNNs that process images, audio, and
  text. We show that media data with provably high \divv\ and \vall\ are produced, and that
  \tool\ achieves good performance w.r.t.~various criteria and significantly
  outperforms all previous frameworks.
\end{itemize}

\begin{tcolorbox}[size=small]
  Full code and data of \tool\ are provided at the website~\cite{snapshot}.
  We will maintain \tool\ to benefit follow-up research comparison and usage.
\end{tcolorbox}

\section{Preliminaries and Motivation}
\label{sec:preliminary}

In this section, we briefly review how different DNNs process perceptions. We
then show key properties of media data and introduce the concept of data
manifold.

\subsection{Deep Neural Network}
\label{subsec:nn}

Most DNNs can be categorized into feed-forward neural networks (FNNs) and
recurrent neural networks (RNNs). We introduce how media data perceptions are
extracted and processed by them.

\parh{FNNs.}~A Convolutional Neural Network (CNN) denotes one
representative FNN. In general, a CNN is trained to focus on the essential
perceptual content of images, which may include object types, colors, and
motions. Its training is mainly accomplished by using the convolutional kernel,
which simulates how humans spot and recognize patterns in images, regardless of
the angle from which the images are seen. FNNs are widely used in image
understanding, e.g., classification, where perceptual content is retrieved for
coarse-grained labeling. Further, perceptual features are extensively tagged
with, for example, driver decision or disease information to facilitate
domain-specific activities such as self-driving and disease diagnostics.
\textit{Translation invariance} is a crucial characteristic enforced by a CNN,
which means that perceptual items (e.g., a cat) may be recognized in an image
even if their appearance changes~\cite{lecun2004learning}.
This characteristic improves a CNN's ability to process a wide range of images
in real-world circumstances. However, as discussed later in
\S~\ref{sec:evaluation}, translation invariance can hamper the efficiency of
previous DNN testing methods that alter images using affine
transformations~\cite{tian2018deeptest}. In contrast, we will show that \tool\
delivers an effective and unified testing pipeline toward FNNs used in general
image understanding and domain-specific tasks (auto-driving and audio
processing) by mutating perceptions.

\parh{RNNs.}~Derived from FNNs, RNNs are capable of processing
variable-length sequences, e.g., natural language sentences. For such data, a
``sequence'' usually contains more informative perceptions than individual
elements do~\cite{mikolov2010recurrent}. RNNs are often used for text
comprehension, such as machine translation (MT), where word dependency in a
language is captured by RNNs. Random mutation can easily damage discrete data
like text; therefore, previous RNN testing relied heavily on predefined
text-mutation templates~\cite{galhotra2017fairness, udeshi2018automated,
ma2020metamorphic, he2020structure}. However, template-based approaches have a
limited generality and are likely to miss faults in real-life corner cases. In
contrast, \tool\ delivers perceptual-level mutation by walking on manifolds
which characterize properties of the language. This enables a thorough
exploration of the RNN input space while simultaneously ensuring the validity of
mutated text.

\parh{DNN Faults.}~A DNN layer can be represented as $\texttt{Act}(W\mathbb{X}
+ b)$, with $\mathbb{X}$ being the input space. $(W,b)$ forms an affine\footnote{
    A geometric transformation that preserves lines and parallelism.
}
transformation over $\mathbb{X}$. The non-linear activate function
$\texttt{Act}$ bends $\mathbb{X}$ to better split regions, e.g., $\texttt{ReLU}:
y = \max(0, x)$ bends the straight line $y = x$ at $(0, 0)$ in a two-dimensional
space. In \F~\ref{fig:func}, a sample neural network solves the XOR problem that
is not linearly separable. The input space is $(x, y) \in [0, 1] \times [0, 1]$
and is labelled as $(x > 0.5) \oplus (y > 0.5)$. By progressively transforming
the input space via each layer, inputs sharing similar properties are
gathered.\footnote{For high-dimensional media data like images, ``inputs sharing
similar properties'' can be dog photos where every dog in the photo is
consistently left-oriented.} Finally, the space is separated (using the
\textcolor{purple}{purple} line in \F~\hyperref[fig:func]{2(c)}) as two linearly
separable regions with different labels. Aligned with existing works,
we characterize DNN faults in the following.

\begin{tcolorbox}[size=small]
DNN faults root from the incorrect separation (i.e., wrong decision
boundaries). Error-triggering inputs in the ill-separated regions have similar
properties, which are typically referred to as DNN \textit{biases}: DNN testing
should be designed to detect such biases.
\end{tcolorbox}

On the other hand, suppose most training data in \F~\hyperref[fig:func]{2(a)}
have $x < 0.5$. The DNN thus primarily relies on the value of $y$ to predict the
XOR output, resulting in \colorbox{pptgreen}{ill-separated regions} in
\F~\hyperref[fig:func]{2(c)}. Error-triggering inputs (\textcolor{pptblue1}{blue
dots}) in these ill-separated regions would have $x > 0.5$ when they are mapped
back to \F~\hyperref[fig:func]{2(a)}. This shows that the faulty DNN has a bias on
the value of $x$. From a holistic view, \tool\ can identify critical perception
changes in real-world media data of different contents and types that expose
\textit{perception biases} of faulty DNNs.

\begin{figure*}[!ht]
    \centering
    % \vspace*{-5pt}
    \includegraphics[width=0.90\linewidth]{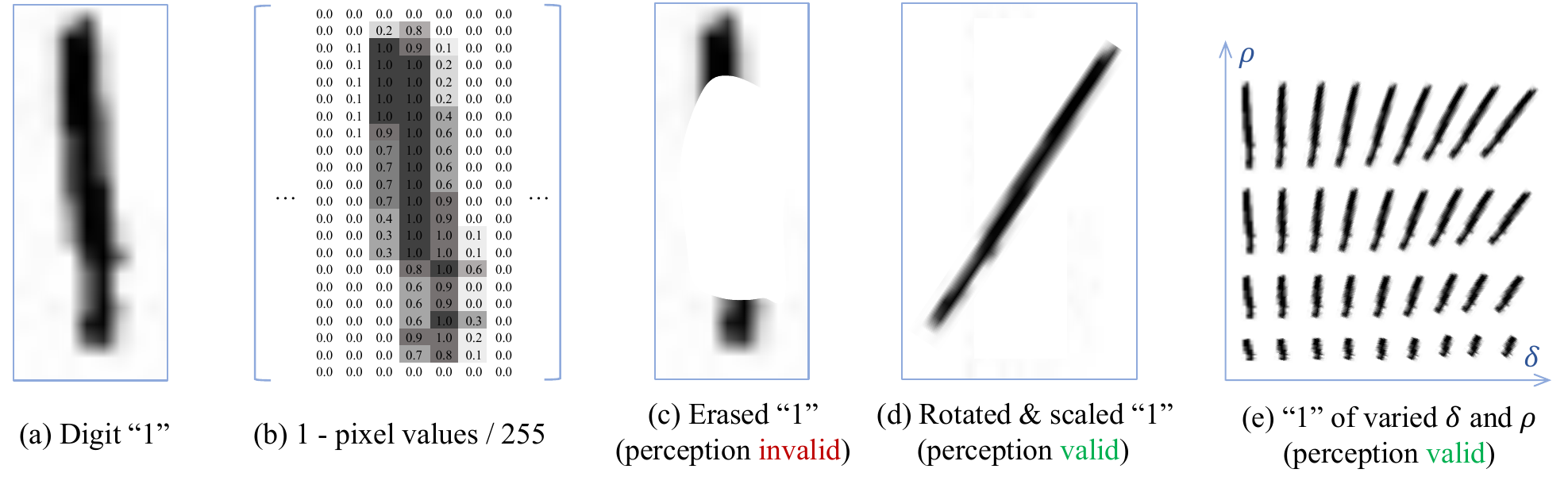}
    %\vspace*{-5pt}
    \caption{Demonstration of key properties of media data using an image of
    digit ``1''. As in (b), to form a valid digit ``1'', there exist constrains
    over pixel values. Comparison between (c) and (d) illustrates the incapability
    of measuring \vall\ using pixel-based criteria. (e) simplifies each digit ``1''
    as a segment and represents them with length $\rho$ and rotation angle $\delta$.
    }
    \vspace*{-5pt}
    \label{fig:mnist}
\end{figure*}

\parh{\mr{DNN Inputs.}}~Also, as shown in \F~\hyperref[fig:func]{2(c)}, the input space is largely
twisted after transformations. Common DNN testing objectives, such as neuron
coverage, aim at maximizing the covered output range of each neuron, which
unavoidably induces invalid inputs (\textcolor{pptred1}{red dots} in
\F~\hyperref[fig:func]{2(c)}) to test DNNs. Though in this demo, we can
explicitly restrict inputs to $[0, 1] \times [0, 1]$ to rule out those invalid
inputs, the constraints formed by perceptual contents over arbitrary real-world
media data bytes are hard to obtain. We therefore point out the following
difficulty of testing DNNs:

\begin{tcolorbox}[size=small]
Given DNNs do not feature ``input checks'' and perceptual constraints over media
data are not explicit, it's demanding albeit challenging to ensure the validity
of mutated media inputs.
\end{tcolorbox}

Previous mutations can frequently break the perceptual contents in media data. 
Recent SOTA works~\cite{kang2020sinvad,dola2021distribution} can generate
perceptually valid images by converting all images into a unified latent space
via generative models. Nevertheless, as will be proved in \S~\ref{sec:formal},
their schemes are inapplicable for real media data, where only meaningless
data are produced.

\subsection{Media Data and Manifold}
\label{subsec:media}

We now elaborate on key properties of media data. Though using images as an example,
mutating other media data suffers from similar challenges introduced in this section.

\F~\hyperref[fig:mnist]{3(a)} shows a $28 \times 28$ image of digit
``1'' (with margins cropped) from the MNIST dataset, which has $784$ dimensions
in the pixel space. As shown in \F~\hyperref[fig:mnist]{3(b)}, pixel values
share perceptual constraints to form a valid ``1'': randomly sampling
pixels from $[0, 255]$ mostly provides noise.

\parh{\mr{Pixel Space.}}~The $784$-dimensional pixel space for digit ``1'' is highly redundant, given
only a small portion of data samples are valid ``1''. Moreover, it is hard to
measure/validate mutations in this space. As shown in \F~\hyperref[fig:mnist]{3(c)},
when most parts of the digit are erased, the mutation breaks the perceptual
contents. In contrast, \F~\hyperref[fig:mnist]{3(d)} rotates and rescales the
digit from \F~\hyperref[fig:mnist]{3(a)}, which still yields a valid ``1''. Note
that previous works often adopt pixel-wise distance criteria to decide the
``validity'' of mutated images, which is improper. For instance, though compared
with \F~\hyperref[fig:mnist]{3(d)}, \F~\hyperref[fig:mnist]{3(c)} has fewer
pixels changed, it is improper to deem \F~\hyperref[fig:mnist]{3(c)} as more
realistic (perceptually valid).

\parh{\mr{Dimension Reduction.}}~To ease the understanding, let's simplify the digit
``1'' as a segment.\footnote{
    A digit ``1'' is more complicated than a segment, but the required
    extra dimensions should not be too many.
} Then, each digit ``1'' can be represented by length $\rho$ and rotation
angle $\delta$ in the polar coordinate, where the dimensions are largely reduced
to two.
\F~\hyperref[fig:mnist]{3(e)} displays digits using $\rho$ and $\delta$.
Obviously, it is easier to form a valid ``1'' by directly restricting the ranges
of $\rho$ and $\delta$. And accordingly, it is also easier to produce perceptually
valid and diverse digits by changing $\rho$ and $\delta$ within the
constrained ranges.

Overall, to enable diverse and valid mutations for arbitrary high-dimensional
media data in real-world settings, it is more desirable to explore a general and
unified dimension reduction strategy for representing media data in a
low-dimensional space that encodes perceptions. 
\tool\ fulfills
the requirements by using data manifold. We now introduce manifold below.

\parh{Manifold.}~Perceptual contents in media data are captured via data
manifold. Overall, manifold forms the basis of \textit{manifold hypothesis},
which states that the high-dimensional space $\mathcal{R}$ of real-world data
embeds the manifold $\mathcal{M}$ of much lower
dimensions~\cite{lee2007nonlinear,fan2019spherereid}. 
For instance, let portrait photos concentrate to one manifold $\mathcal{M}_p$,
$\mathcal{M}_p$ shall retain constraints over certain pixels that jointly make
up perceptions like head, ear and nose. 
Generally, manifolds preserve the ``closeness'' in perceptions of
high-dimensional data. Also, data of the same class lie in one continuous
manifold, whereas data of distinct classes (e.g., car vs.~cat) lie in
\textit{disconnected} manifolds~\cite{lee2007nonlinear,fan2019spherereid}.
Manifolds are typically adopted for dimensionality reduction, e.g., for
PCA~\cite{abdi2010principal}, where the principal components of each data point
$x$ should uniquely locate its projection on
$\mathcal{M}$~\cite{bengio2013representation}.

\noindent \textbf{Manifold Construction.}~Manifolds used for PCA are constructed
via linear functions. Nevertheless, linear approaches are not expressive since
most real-world mappings between $\mathcal{M}$ and $\mathcal{R}$ are non-linear.
Several non-linear methods are thus
proposed~\cite{balasubramanian2002isomap,roweis2000nonlinear,belkin2003laplacian}.
These approaches are non-parametric, which only enable low-dimensional
representations and are incapable of inferring other data points lying on
$\mathcal{M}$. This work relies on manifold to mutate media data of different
types in a provably diverse (\divv) and valid (\vall) manner for DNN testing.
Enabled by recent advances in generative models~\cite{zhu2016generative}, we
construct projections $f_{\theta}: \mathcal{R} \rightarrow \mathcal{M}$ using
\textit{parametric} and \textit{non-linear} approach, i.e., carefully crafted
generative models (which are different from those in SOTA works;
see details in \S~\ref{sec:design}).

\begin{figure}[!ht]
    %\vspace{-10pt}
    \centering \includegraphics[width=1.01\linewidth]{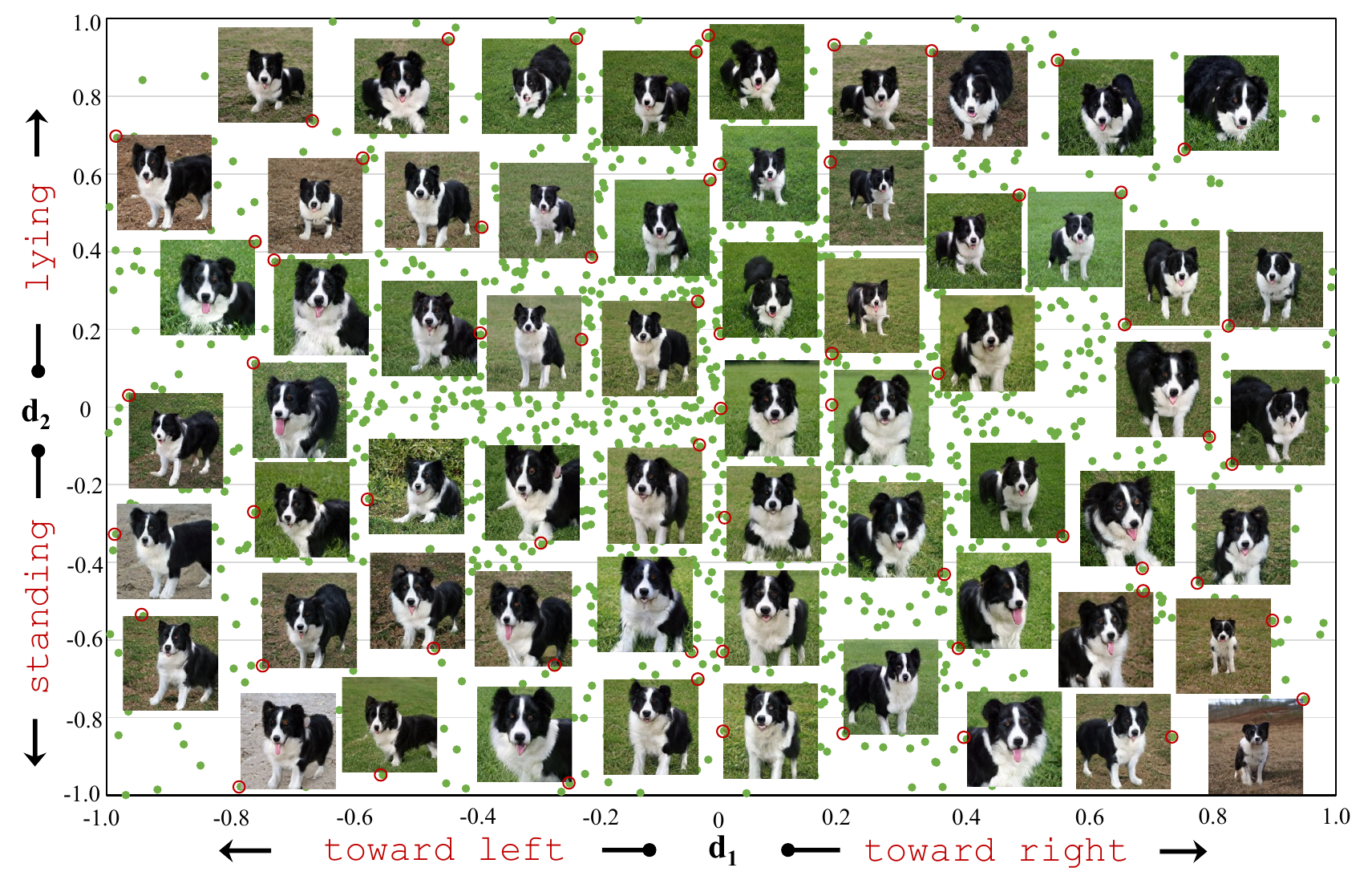}
    \vspace{-10pt}
    \caption{Visualization of the manifold (constructed by \tool) of Border Collie
    photos, where Border Collies sharing similar properties are gathered (e.g., 
    left-oriented when $d_1 < 0$).}
    \label{fig:visualize}
    \vspace{-10pt}
\end{figure}

\noindent \textbf{\mr{Image Manifold.}}~\F~\ref{fig:visualize} depicts the variety
of Border Collie photos on a manifold $\mathcal{M}$. Using \tool, we project a
set of Border Collie photos into coordinate $z$ on $\mathcal{M}$ of 120
dimensions. To ease the presentation, we further pick the first two $z$-values,
$d_1$ and $d_2$, to form a 2D manifold on \F~\ref{fig:visualize}. We then pick
coordinates on \F~\ref{fig:visualize}, marked in \textcolor{pptred}{red cycles},
and use \tool\ to map them to images.
Despite its two-dimensionality, the visualization in \F~\ref{fig:visualize} is
informative to clarify the concept of manifold. The hue of the background grass
gathers images. Also, most Collies stand when $d_2<0$ and lie down if $d_2 > 0$
(particularly when $d_1 = 0$). The value of $d_1$ seems to be related with the
orientation since most of the Collies look left when $d_1 < 0$, and vice versa.

\parh{\mr{Text/Audio Manifold.}}~Manifolds of text data constrain primarily the word dependencies, which further
encode the semantics and grammatical coherence of
sentences~\cite{cai2011manifold,guo2020nonlinear}. For instance, mutating
sentences in $\mathcal{M}$ can modify the subjects or change the described
events (see \F~\ref{fig:demo}). Similarly, audio manifolds may enclose the
spoken words and tones~\cite{li2017unsupervised,briggs2009audio}, allowing for
changes in intonation when mutating audios via manifolds.

We clarify that manifold is more \textit{comprehensive} than training data.
For \F~\ref{fig:visualize}, if we only have images whose projected $d_1$ and $d_2$
have the same sign, i.e., \texttt{$\langle$lying, toward right$\rangle$} Collies and
\texttt{$\langle$standing, toward left$\rangle$} Collies, by inferring
perceptual-level changes from these images, manifold $\mathcal{R}$ in
\F~\ref{fig:visualize} can still be created and enables the construction of
\texttt{$\langle$lying, toward left$\rangle$} Collies and
\texttt{$\langle$standing, toward right$\rangle$} Collies. This reveals the
feasibility to launch full-fledged DNN testing by systematically exploring
manifolds.

\noindent \textbf{Walking on a Manifold.}~Performing perceptual-level mutations
over media data involves two orthogonal aspects, namely, mutating which
perceptions and to what extent a perception is mutated. These two aspects can be
characterized as the \textit{walking direction} and the \textit{step size} on a
manifold. Suppose the training data corresponding to \F~\ref{fig:visualize} is
biased to standing Collies such that the trained DNN incorrectly relies on
``standing'' to classify an image as Collies. \tool, by walking upwards with a
large step size, mutates Collies to be lying down, effectively stressing DNNs
and likely triggering mis-classifications. We present several examples of
perceptual-level sentence/image mutation in \F~\ref{fig:demo}. The mutation is
performed when \tool\ walks towards one direction on the manifold and each
sentence/image denotes one footprint of \tool. 

\begin{figure*}[!ht]
  \centering
  \includegraphics[width=0.85\linewidth]{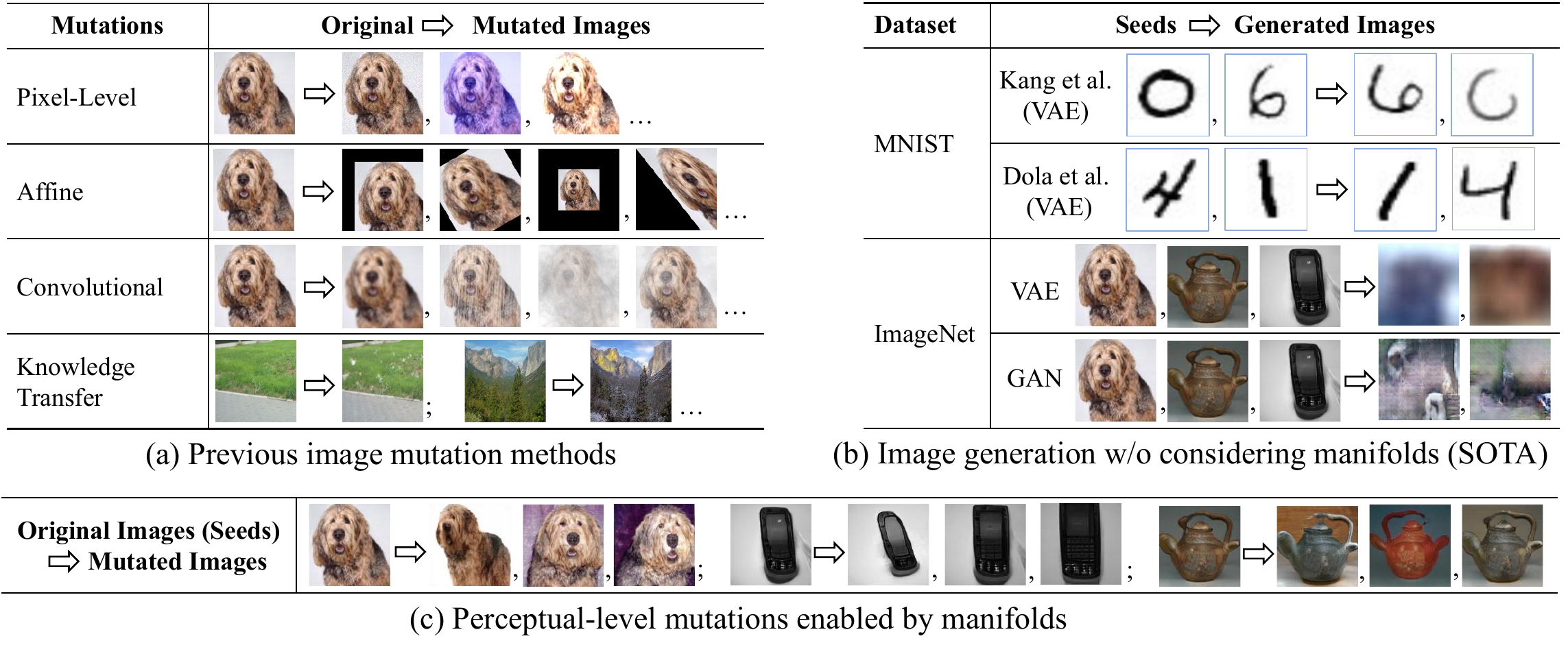}
  \caption{Illustration of previous (a) image mutations and (b) image generation
  w/o consider manifolds. (c) shows the perceptual-level mutations of \tool\
  enabled by data manifold.}
  \label{fig:table}
\end{figure*}

\section{Related Work}
\label{sec:motivation}

We review existing works in generating diverse (\divv) media data in
\S~\ref{subsec:motivation-mutation} and \S~\ref{subsec:motivation-generation}.
We then review methods to preserve format and perception validity (\vall) of
mutated media data in \S~\ref{subsec:motivation-validation}.

\subsection{Producing Diverse Media Data with Mutation}
\label{subsec:motivation-mutation}

\F~\ref{fig:table} compares image mutation launched by existing methods and by
\tool. We summarize the application of existing mutation methods to images and
discuss their extension to text:

\noindent \underline{Pixel-Level Transformation}~changes image brightness or
contrast, or adds noise. DeepXplore~\cite{pei2017deepxplore} uses pixel-wise mutations
in a whitebox context under the guidance of target DNN's gradients. It comprises
two schemes: randomly adding black blocks or perturbing user-specified regions
with noise generated from gradients.
Pixel-level mutations can be ineffective as images
usually have many pixels. Both approaches may damage the image \vall,
and even violate labels of original images (e.g., tearing objects in a photo).

\noindent \underline{Affine Transformations}~translate/rotate/scale/shear an image
while preserving the collinearity after mutation (see \F~\hyperref[fig:table]{5(a)}).
As discussed in \S~\ref{subsec:nn}, the \textit{translation invariance} of a CNN
may hinder affine transformations.

\noindent \underline{Convolutional Transformations} perform
holistic modification of an image, such as blurring or adding domain-specific
(e.g., rainy) filters.

Affine and convolutional transformations are first considered
in DeepTest~\cite{tian2018deeptest}. DeepSmart~\cite{demir2019deepsmartfuzzer}
divides an image into sections and mutates one at a time. Modifying only one
section may result in fragmented images.
Overall, despite affine/convolutional transformations offer more
holistic mutations, these changes are usually limited and require
domain knowledge (e.g., a traffic scene should be rotated by a
small angle). Thus, the ``behaviors'' of target DNNs are often not fully
tested.

\noindent \underline{Knowledge Transfer}~transforms data from one domain to
another by leveraging knowledge from the target domain.
DeepRoad~\cite{zhang2018deeproad} and TACTIC~\cite{li2021tactic} mutate images
using \textit{style transfer} techniques~\cite{zhu2017unpaired,liu2017unsupervised}.
Their strength (and limits) stems from \textit{cycle-consistency}
training~\cite{zhu2017unpaired}, a popular learning paradigm for generative
models. For example, given street images taken on sunny and snowy days, these
techniques enhance sunny-street images by gradually changing the weather to
snowy, and vice versa. Weather conditions provided by ``cycle-consistency'' are
more realistic than those added by convolutional transformations and
are not confined to a few available templates. However, the availability of
photos from two domains limits this strategy (e.g., photos taken from two
weather conditions are needed). ``Knowledge transfer'' also fails when images in
two domains have significant geometric distinctions~\cite{Cycleganfail}. This
method has only been used to test autonomous driving~\cite{zhang2018deeproad}.
It is difficult to extend the method to arbitrary real-life scenarios.

\noindent \textbf{Text.}~Existing works heavily study mutating images. However,
real-world DNN models analyze various types of media data, such as text. A
``pixel-level'' mutation similar to that for images is impractical in the case
of text because random perturbation of words will not retain grammatical or
linguistic coherence. Most text mutations are performed using pre-defined
templates or heuristics~\cite{galhotra2017fairness, udeshi2018automated,
  ma2020metamorphic, he2020structure}.

\subsection{Ensuring Format and Perception Validity of Mutated Media Data}
\label{subsec:motivation-validation}

On top of the restrictions defined by formats (e.g., RGB images' pixel values
must in $[0, 255]$), the \vall\ of media data requires perceptual contents being
retained after mutations (e.g., a cat is still a ``cat''). To clarify, \vall\ is
often referred to as \textit{realism} of mutated media data by previous
works~\cite{xie2018coverage,tian2018deeptest,zhang2018deeproad}.
Mutations introduced in \S~\ref{subsec:motivation-mutation} can generate
unrealistic data, and therefore, extra validation procedures are performed by
existing works. A grammatical check can validate natural language sentences.
However, validating mutated images is more difficult and often involves
heuristics.

TensorFuzz~\cite{odena2018tensorfuzz} adds pixel-wise noise to images. It
enforces the realism of mutated images by truncating accumulated noise to a
user-defined range. DeepTest~\cite{tian2018deeptest} limits the parameters of
each transformation to a specific range. DeepTest also assesses mean squared
error (MSE) to rule out likely broken images. However, as found in
\S~\ref{subsec:effectiveness}, many of its mutated images, though passed its
validation, are barely identifiable to humans.
DeepHunter~\cite{xie2018coverage} allows only one-time affine transformation on
each image. It uses thresholds to bound \#mutated pixels and mutated pixel
values for pixel-level transformations. 
However, as discussed in \S~\ref{subsec:media}, pixel-based criteria can
hardly ensure \vall, let alone the criteria itself is sensitive to thresholds.
DeepRoad~\cite{zhang2018deeproad} ensures authenticity by rejecting an image
with features that are significantly different from features in training data.
The features are extracted by a pre-trained oracle model
(VGG~\cite{simonyan2014very}). It also assumes the availability of the target
DNN's training dataset, which may not be available.

Moreover, as shown in \S~\ref{subsec:effectiveness}, since pixel-level mutations
directly change pixel values, existing validation schemes can neglect images of
invalid format (e.g., pixel values outside $[0, 255]$) if the format restriction
is not explicitly considered. They also frequently overlook images of broken
perceptions. \tool\ only mutates perceptions and always produces format-valid
media data (as it never directly alters pixel values). The perceptual contents
are also meaningful under the constraints formed by data manifold;
see \S~\ref{subsec:tradeoff} for details.

\subsection{Generating Diverse \& Valid Images Using Toy Datasets with Generative Models}
\label{subsec:motivation-generation}

Recent
works~\cite{kang2020sinvad,dola2021distribution,byun2020manifold,byun2020manifolda,fuzzgan},
including SOTA works~\cite{kang2020sinvad,dola2021distribution}, use generative
models like generative adversarial network (GAN)~\cite{goodfellow2014generative}
and variational auto encoder (VAE)~\cite{kingma2014auto} to generate test inputs.
Holistically speaking, a generative model is trained to map a seed dataset corpus
to a continuous distribution, a.k.a., a unified latent space. Once
trained, new inputs can be generated by sampling from the latent space. Since
data are mapped to a continuous distribution, infinite samples can be generated
as test inputs.

Note that these SOTA works are not effective in generating media inputs with
high \divv\ and \vall. Kang et al.~\cite{kang2020sinvad} focuses on
generating images that ``lie on the semantic \textit{boundary} of labels.'' They
thus generate media data with obscurely defined labels (e.g., a fused digit of
``4'' and ``9''), which is less useful for DNN testing when checking the prediction
consistency. In \F~\hyperref[fig:table]{5(b)}, we present some examples
generated by~\cite{kang2020sinvad}, whose key limitation is also pointed out
by~\cite{dola2021distribution}. Instead, Dola et al.~\cite{dola2021distribution}
samples media data ``within distributions.''
However, as will be proved in \S~\ref{subsec:validity}, these SOTA works are
only applicable to toy datasets like MNIST, and \vall\ is deemed to be nullified
when their methods are used for real-world images. We present examples in
\F~\hyperref[fig:table]{5(b)}: when following their strategies to mutate images
from ImageNet~\cite{imagenet}, only meaningless, broken images are generated. In
contrast, \tool\ can generate meaningful and diverse images using ImageNet as
the seed corpus, as shown in \F~\hyperref[fig:table]{5(c)}.

\section{Formal Analysis}
\label{sec:formal}

In this section, we formally define \vall\ and \divv\ of a test suite based on
manifold and show how they inextricably bound each other. We then demonstrate
how to provably increase \vall\ and \divv.

\subsection{\gen: A Unified Formulation of Media Input Mutation/Generation Methods}
\label{subsec:data-generation}

We first formulate the input mutation (\S~\ref{subsec:motivation-mutation}) and
input generation (based on generative models;
\S~\ref{subsec:motivation-generation}) methods of existing testing works as a
unified data generation (\gen) process. 

% \vspace{-5pt}
\begin{definition}[\gen]
  \label{def:generation}
  Given media data $x$, mutating $x$ to $\hat{x}$, or generating $\hat{x}$ from
  $x$, can be formulated as a unified data generation process $\hat{x} =
  G_{\theta_2}(E_{\theta_1}(x), h)$. $E_{\theta_1}$ is the encoder converting
  $x$ into a (latent) representation. $G_{\theta_2}$ is the generator and
  $h \in H$ specifies how $\hat{x}$ is different from $x$. $\theta_1, \theta_2 \in
  \emptyset \cup \Theta$ and $\Theta$ is the parameter space.
\end{definition}
% \vspace{-5pt}

$\theta_1$ and $\theta_2$ can be $\emptyset$, and therefore, most mutation-based
\gen\ introduced in \S~\ref{subsec:motivation-mutation} can be written as a
non-parametric form $G(E(x), h)$, where $E(x) = x$, $h = [a, b]^T \in H$ and
$G(E(x), h) = [x, 1] \times h = ax + b$. Concretely, for byte-level mutations,
$a$ and $b$ are floating numbers. For example, to make a picture brighter,
$a=1.0$ and $b > 0$. For affine/convolutional transformations, $a$ and $b$ should
be matrices/vectors, e.g., $a$ is the rotation matrix and $b$ is zero vector when
rotating $x$. 
\gen\ based on style transfer can be represented as $G_{\theta_2}(E(x), h)$,
where $\theta_2 \in \Theta$ and $E(x) = x$. $G_{\theta_2}$ is the style
translator, and $h \in H$ is the parameter that decides which and to what extent a
style is transferred over $x$.

For SOTA generative model-based \gen, $\forall x \in \mathbb{X}, E_{\theta_1}(x)
\equiv \mathbb{Z}$, where $\mathbb{X}$ is the pixel space and $\mathbb{Z}$ is
the low-dimensional representation of $\mathbb{X}$, e.g., the unified latent
space of~\cite{kang2020sinvad}, or the ``distribution'' referred
in~\cite{dola2021distribution}. Then, $G_{\theta_2}(\mathbb{Z}, h) =
g_{\theta_2}(z \overset{h}{\sim} \mathbb{Z})$, where $g_{\theta_2}$ is the
generator trained over images from different manifolds and $z \overset{h}{\sim}
\mathbb{Z}$ denotes sampling $z$ from $\mathbb{Z}$ according to criterion $h$.
Concretely, $h$ in~\cite{kang2020sinvad} aims to sample images ``lying in
distribution boundaries'' whereas $h$ in~\cite{dola2021distribution} samples
``in-distribution'' images.

In our work, \tool, $\hat{x} = G_{\theta_2}(E_{\theta_1}(x), h) =
\mathcal{G}_{\theta_2} (\mathcal{E}_{\theta_1}(x) \times h)$, where
$\mathcal{E}_{\theta_1}$ is the encoder that converts $x$ to its representation
in the manifold and $\mathcal{G}_{\theta_2}$ is the generator that maps from
each manifold $\mathcal{M} \subset \mathbb{M}$ to its corresponding region in
$\mathbb{X}$. Both $\mathcal{E}_{\theta_1}$ and $\mathcal{G}_{\theta_2}$
consider the separation between manifolds. $h$ is the perceptual transformation
launched on each $\mathcal{M} \subset \mathbb{M}$ that mutates certain
perceptions under feedback of some off-the-shelf DNN testing criteria.
Based on \Df~\ref{def:generation}, we have the following two assumptions.

% \vspace{-5pt}
\begin{assumption}[Gaussian Input]
  \label{asmp:Gaussian}
  The input of $G_{\theta \in \emptyset \cup \Theta}$ in \gen\
  (\Df~\ref{def:generation}) is sampled from a Gaussian distribution.
\end{assumption}
% \vspace{-5pt}

\begin{proof}[Analysis of \Asmp~\ref{asmp:Gaussian}]
  Inputs of generative models, by design, are sampled from either uniform or normal
  distributions (i.e., a special case of Gaussian distribution). Because uniform
  distribution can be converted from Gaussian distribution~\cite{ross2010first},
  \Thm~\ref{asmp:Gaussian} holds for generative model-based \gen.
  
  For mutation-based \gen, given an input $x$, all possible mutated images
  are $\hat{\mathbb{X}}_{x} \leftarrow [x, 1] \times H$.\footnote{
  $\hat{\mathbb{X}}_{x} \leftarrow G_{\theta_2}(x, H)$ for input mutation based
  on style transfer. In this section, we primarily discuss non-parametric input
  mutations whose conclusion can be directly extended to style transfer-based
  mutations.} To our best knowledge, all prior works sample $h$ based on either
  normal distribution or uniform distribution.\footnote{In practice, the de
  facto image library, OpenCV~\cite{bradski2000opencv}, only provides sampling
  pseudo random numbers from normal or uniform distributions. OpenCV is
  extensively used by most, if not all, DNN testing research.} For normal distribution,
  $h \sim \mathcal{N}(H_{\min } +
  \frac{H_{\Delta}}{2}, (I \cdot H_{\Delta})^2) =
   H_{\Delta} \cdot \mathcal{N} (H_{\min} + \frac{H_{\Delta}}{2}, I^2)$ where
   $H_{\Delta} = H_{\max} - H_{\min}$ and $I$
  is the identity matrix. Therefore, $\hat{\mathbb{X}}_{x} \leftarrow [x, 1]
  \times z  H_{\Delta}$ where $z \sim \mathcal{N} (H_{\min} + \frac{H_{\Delta}}{2}, I^2)$.
  The same applies if $h \sim H$ is based on uniform
  distribution. Therefore, for a given $x$ and $H$, $\hat{\mathbb{X}}_{x}$ is
  generated by taking inputs $z$ sampled from Gaussian distribution;
  \Asmp~\ref{asmp:Gaussian} thus holds.
  In \tool, as will be shown in \S~\ref{sec:design}, since $h$ is a covariance
  matrix and $\mathcal{E}_{\theta_1}(x)$ follows normal distribution, inputs to
  $\mathcal{G}_{\theta_2}$ are accordingly samples from Gaussian distribution.
\end{proof}

% \vspace{-5pt}
\begin{assumption}[Lipschitz Constant]
\label{asmp:lip}
  $\exists L > 0$, the Lipschitz constant of $G_{\theta \in \emptyset \cup \Theta}$
  is no larger than $L$, i.e., $\forall (z_i, z_j)$ satisfy
  $$
  \sup_{z_i \neq z_j} \frac{||G_{\theta}(z_i) - G_{\theta}(z_j)||}{||z_i - z_j||} \leq L
  $$
\end{assumption}
% \vspace{-5pt}

\begin{proof}[Analysis of \Asmp~\ref{asmp:lip}]
  \label{pf:lip}
  For non-parametric \gen, as discussed in Analysis of \Asmp~\ref{asmp:Gaussian},
  since $|G_{\theta}(z_i) - G_{\theta}(z_j)| / |z_i - z_j|$ is a constant,
  \Asmp~\ref{asmp:lip} is therefore always true.
  
  For generative model-based \gen, \Asmp~\ref{asmp:lip} should also hold since,
  conceptually, well-trained generative models are leveraged for \gen, whose
  Lipschitz constants are bounded. Otherwise, $|G_{\theta}(z_i) - G_{\theta}(z_j)| / |z_i - z_j| \rightarrow \infty$,
  such that the generated images are corrupted --- the generative models
  can hardly generate useful test inputs~\cite{arjovsky2017wasserstein,miyato2018spectral}.
  On the other hand, in practice, various approaches for bounding the Lipschitz constant
  are adopted when training generative models, such as weight clipping, spectral normalization,
  etc~\cite{arjovsky2017wasserstein,miyato2018spectral,zhang2019self,brock2018biggan}.
  These approaches support \Asmp~\ref{asmp:lip} from a technical perspective.

\end{proof}

\subsection{Quantifying Perception \vall\ and \divv\ of Media Data Generated via \gen}
\label{subsec:validity}

\Df~\ref{def:generation} cohesively formulates existing media data
mutation/generation methods as \gen. Inspired by the precision \& recall
criteria proposed by~\cite{sajjadi2018assessing} for measuring generative
models, we formulate \vall\ and \divv. Let the seed corpus be $\mathcal{S}$,
which lie on a collection of manifolds $\mathcal{M} \subset \mathbb{M}$. Let the
data generated (according to \Df~\ref{def:generation}) via $G_{\theta}$ from
$\mathcal{S}$ be $\hat{\mathcal{S}}_{\theta}$. Intuitively, \vall\ of
$\hat{\mathcal{S}}_{\theta}$ can be quantified as the proportion of
$\hat{\mathcal{S}}_{\theta}$ that lie in $\mathbb{M}$. Accordingly, \divv\ of
$\hat{\mathcal{S}}_{\theta}$ can be quantified as the proportion of $\mathcal{M}
\subset \mathbb{M}$ that is covered by $\hat{\mathcal{S}}_{\theta}$. Formally, we
have the following definition.

% \vspace{-5pt}
\begin{definition}[\vall\ and \divv]
  \label{def:vd}
  Given corpus $\mathcal{S}$ lying on manifolds $\mathbb{M}$ with associated
  distribution $\mathcal{P}$ (note that as introduced in \S~\ref{subsec:media},
  $\mathcal{P}$ is discontinuous if multiple manifolds exist). Suppose test suite
  $\hat{\mathcal{S}}_{\theta}$ generated (see \Df~\ref{def:generation}) over
  $\mathcal{S}$ has distribution $\mathcal{Q}$. Let $\Lambda =
  {supp}(\mathcal{P}) \cap {supp}(\mathcal{Q})$, where ${supp}(\mathcal{P})$ is
  the support of $\mathcal{P}$.\footnote{For the pixel space $\mathbb{X}$,
  ${supp}(\mathcal{P}) = \{x \in \mathbb{X}|\mathcal{P}(x) > 0\}$.} Then,
  $\mathcal{P}$ is the mixture of $\mathcal{P}_{\Lambda}$ and
  $\mathcal{P}_{\overline{\Lambda}}$ which are distributions defined on
  $\Lambda$ and the complement of $\Lambda$, respectively. Similarly,
  $\mathcal{Q}$ can be defined over $\mathcal{Q}_{\Lambda}$ and $\mathcal{Q}_{\overline{\Lambda}}$.
\end{definition}

% \vspace{-5pt}
\begin{subdefinition}[\vall\ $\alpha$]
  \label{subdef:validity}
  $\hat{\mathcal{S}}_{\theta}$ has a validity score $\alpha \in [0, 1]$ such that
  $\mathcal{Q} = \alpha \mathcal{Q}_{\Lambda} + (1 - \alpha)\mathcal{Q}_{\overline{\Lambda}}$.
\end{subdefinition}
% \vspace{-5pt}

\begin{subdefinition}[\divv\ $\beta$]
  \label{subdef:diversity}
  $\hat{\mathcal{S}}_{\theta}$ has a diversity score $\beta \in [0, 1]$ such that
  $\mathcal{P} = \beta \mathcal{P}_{\Lambda} + (1 - \beta)\mathcal{P}_{\overline{\Lambda}}$.
\end{subdefinition}
% \vspace{-5pt}

\parh{Validity vs. Diversity}~Based on theoretical results
of~\cite{tanielian2020learning}, we have the following important Lemma.

\begin{lemma}[The None Land]\footnote{\Lem~\ref{lem:none} is originally
  referred as ``no GAN's land'' in~\cite{tanielian2020learning}. Nevertheless, its
  theory is not limited to GAN (it only requires $G$ satisfying
  \Asmp~\ref{asmp:Gaussian} and \Asmp~\ref{asmp:lip}) and we rename it as
  ``the None Land'' to better reflect our view.}
  \label{lem:none}
  For a generator $G_\theta$ that satisfies \Asmp~\ref{asmp:Gaussian} and
  \Asmp~\ref{asmp:lip}, suppose its generated test suite $\hat{\mathcal{S}}_\theta$
  captures $\hat{K}$ manifolds, which are $\hat{\alpha}$ overlapped with the
  manifold collection $\mathbb{M}$. Let $\mathbb{M}$ has total $K$ disconnected
  manifolds, where $\hat{\beta}$ percentage is covered by $\hat{\mathcal{S}}_\theta$.
  Thus, $\hat{K} \geq K\hat{\beta}$ and
  \begin{equation*}
    \begin{aligned}
      \hat{\alpha} &\leq
      \exp \left(\frac{1}{2}(\frac{D}{2L})^2 + \frac{D}{2L}\sqrt{2\log\hat{K}}\right)^{-1} \\
      &\leq 
      \exp \left(\frac{D}{2L}\sqrt{2\log K\hat{\beta}}\right)^{-1}
    \end{aligned}
  \end{equation*}
  
  \noindent holds, where $D$ is the minimal distance between any two manifolds and $L$ is
  from \Asmp~\ref{asmp:lip}. Importantly, when $K$ is large (e.g., $K=1,000$ for
  real-world media datasets like ImageNet), $\hat{\alpha}$ is negligible, with 
  induced outputs provably falling into the ``None'' region (e.g.,
  generated images are meaningless).
\end{lemma}

\parh{Interpretation.}~$\hat{\alpha}$ and $\hat{\beta}$ quantify perception
\vall\ and perception \divv\ defined in \Df~\ref{def:vd}, respectively. It
explains why SOTA methods can succeed on a toy dataset like MNIST by generating
diverse images of convincing realism. MNIST has only 10 classes, corresponding
to ten manifolds (i.e., $\hat{K} = 10$). Moreover, $D$, the minimal distance of
two manifolds (e.g., manifolds for digits ``4'' and ``9''), is small since every
MNIST image comprises a white digit on a black background. 

However, we point out that for real-world media datasets like ImageNet, $D$ is
often large. For example, what and how perceptions form a car are distinct with
that of a dog. Also, with 1000 classes in ImageNet, $\hat{K}$ is large (i.e., 1000)
and $\alpha$ is suppressed to negligible, leading to ``None'' outputs.
In sum, \Lem~\ref{lem:none} illustrates a critical albeit overlooked limit of
SOTA works:

% \vspace{-5pt}
\begin{tcolorbox}[size=small]
The SOTA methods employ generative models for \gen, which captures full
manifolds of corpus $\mathcal{S}$ in a unified distribution. However, when $D$
and $\hat{K}$ are large, \vall\ of generated data $\hat{\mathcal{S}}_{\theta}$ is
inevitably small, resulting in outputs falling into the ``None'' region. That is,
when $D$ and $\hat{K}$ increase, samples generated by SOTA works would fall
outside of $\mathbb{M}$ and are meaningless.
\end{tcolorbox}
% \vspace{-5pt}

Having that stated, previous non-parametric \gen\ does not suffer from
this problem since $\forall x \in \mathbb{X}, E(x) = x$; the transformed
Gaussian inputs (see \Df~\ref{def:generation}) are separated by each input. 

Viewing their pros and cons through the lens of manifold, our intuition is to
explore the conceptual synergy of existing \gen\ by \textit{treating individual
manifold separately}. This way, we limit $K$ to a small value ($K$ may be
slightly over one since there may be estimation errors) and overcome the ``The
None Land'' problem. 
\Lem~\ref{lem:none} also shows that given $D$ and $K$ derived from the seed
corpus $\mathcal{S}$, $\alpha$ and $\beta$ inextricably bound each other, though
the bound is \textit{not} tight. For instance, when $K$ is close to 1, the bound
approximates 1. This reveals opportunities to provably enhance \divv\ and \vall\
when generating media data, as will be discussed in \S~\ref{subsec:tradeoff}.

\subsection{Retaining \vall: A Mechanical Way}
\label{subsec:tradeoff}

\begin{figure}[!ht]
  % \vspace{-10pt}
  \centering \includegraphics[width=0.8\linewidth]{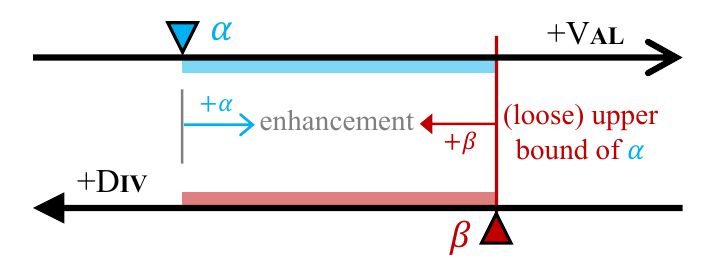}
  % \vspace{-15pt}
  \vspace{-5pt}
  \caption{Improving \vall\ and \divv\ due to the loose bound.}
  \label{fig:tradeoff}
  % \vspace{-10pt}
\end{figure}

\begin{figure*}[!ht]
  \centering
  % \vspace{-20pt}
  \includegraphics[width=0.8\linewidth]{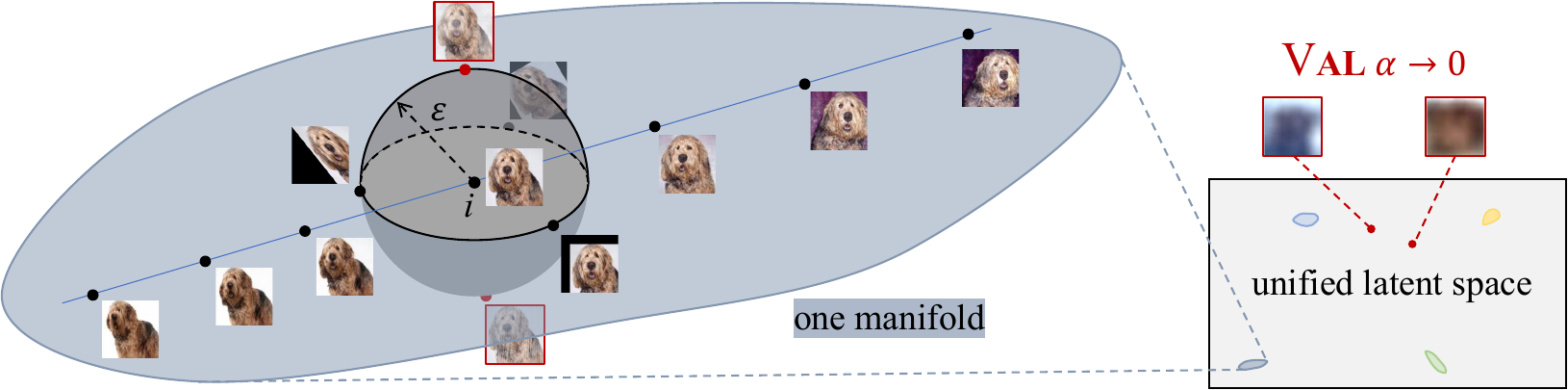}
%  \vspace{-5pt}
  \caption{Comparison between \tool\ and existing works by projecting their
    mutated images on manifold. Previous mutation approaches (left) can hardly
    mutate the perceptions. Similarly, existing input generation schemes (right),
    which do not consider the separation among manifolds, only generate meaningless
    inputs when applied on real datasets.}
  \label{fig:manifold}
  % \vspace{-15pt}
\end{figure*}

\Lem~\ref{lem:none} shows $\alpha$ and $\beta$ bound each other. Therefore,
increasing \divv\ would reduce the upper bound of \vall, and vice versa.
Nevertheless, we clarify that \vall\ can be retained to a reasonably high
extent, if not improved, when increasing \divv. The reasons are two-fold. First,
while $\beta$ upper bounds $\alpha$ according to the ``The None Land'' theory, we
find that the bound is loose in practice. For instance, when $K$ is
small (e.g., close to 1), the bound approximates 1, according to this theory.
Thus, as seen in \F~\ref{fig:tradeoff}, $\alpha$ and $\beta$ can be enhanced
simultaneously. Second, since manifolds preserve the closeness among data points
(see \S~\ref{subsec:media}), the region in $G_{\theta}$'s input space where the
closeness is likely violated is possibly out-of-manifold, denoting inputs
breaching \vall\ and having no contribution to \divv. Therefore, we propose a mechanical way to retain high \vall, by
identifying and pruning samples in $\hat{\mathcal{S}}_{\theta}$ that have high
local sensitivity. We first characterize the local sensitivity of a \gen's
input.

% \vspace{-3pt}
\begin{definition}[Sensitivity]
  \label{def:sensitivity}
  For an arbitrary input $z$ of $G_{\theta}$, the Jacobian matrix
  $J_{G_\theta}$ of the generator ${G_\theta}$ with respect to $z$ is
  defined as:
  $$J_{G_\theta}(z)_{i,j} = \frac{\partial G_\theta (z)_i}{\partial z_j},$$
  where $J_{G_\theta}(z)_{i,j}$ is the $(i,j)$-th entry of $J_{G_\theta}(z)$.
  Accordingly, the local sensitivity of $z$ towards $G_{\theta}$ is characterized
  by the Jacobian Frobenius Norm (JFN):
  $$||J_{G_\theta}(z)||_F^2 = \sum\nolimits_{i} \sum\nolimits_{j}
  \left(J_{G_\theta}(z)_{i,j}\right)^2$$
\end{definition}

The intuition is that, the local sensitivity around $z$ can be characterized via
its gradient~\cite{novak2018sensitivity}, which describes how much the output
changes w.r.t.~input perturbations. Considering previous media data mutation
methods, where each $x \in \mathbb{X}$ is formulated as
$\hat{\mathbb{X}}_{x} \leftarrow [x, 1] \times z H_{\Delta}$ and $z$
is regarded as the Gaussian input to $G_{\emptyset}$, the local sensitivity of
$z$, by following JFN, is thus only (positively) decided by
$H_{\Delta} = H_{\max} - H_{\min}$.
Furthermore, according to the theoretical and empirical results
in~\cite{tanielian2020learning,brock2018biggan}, we have the following lemma:

% \vspace{-3pt}
\begin{lemma}[Out-of-Manifold]
  \label{lem:oom}
    Let a collection of data $S$ generated by ${G_\theta}$ has the \vall\ score as 
    $\alpha$. Let $|\cdot|$ denote the size of a set. We select a subset $S_1$
    with the highest sensitivity from $S$, where $|S_1| = \tau |S|$
    and $\tau < 1 - \alpha$. Suppose the out-of-manifold data
    in $S_1$ forms a set $S_2 = \{x \; | \; x \in S_1, x \notin \mathbb{M}\}$,
    we have:
    $$ 
       \tau(1 - \alpha) < \frac{|S_2|}{|S|}
    $$
\end{lemma}

\Lem~\ref{lem:oom} demonstrates that when selecting data generated by 
\gen\ with the highest sensitivity, the out-of-manifold data are more likely to be
chosen. Therefore, by pruning these data, the \vall\ score $\alpha$ should be
increased. Concretely, we present the following theorem:

% \vspace{-3pt}
\begin{theorem}[Increasing \vall\ when Retaining \divv]
  \label{thm:control}
    Given a collection of data samples $S$ generated by ${G_\theta}$ whose \vall\ score 
    is $\alpha$ and \divv\ score is $\beta$, we prune a proportion of $\tau$
    samples, dubbed as $S_1$, that have the highest sensitivity, where
    $\tau < 1 - \alpha$. Let $p|S|$ be the number of out-of-manifold
    samples in the pruned data (i.e., $|S_2|$ in \Lem~\ref{lem:oom}), then we can
    compute the new \vall\ score $\alpha^\uparrow$ and \divv\ score $\beta^\approx$
    as follows: $$\alpha^\uparrow = \frac{\alpha - \tau + p}{1 - \tau}, \qquad
    \beta^\approx = \frac{\alpha - \tau + p}{\alpha} \beta.$$
    \Lem~\ref{lem:oom} trivially provides $p > \tau(1 - \alpha)$, and
    as a consequence, we have $$\alpha^\uparrow > \alpha \qquad \text{and} \qquad
    \beta^\approx > (1 - \tau) \beta$$
\end{theorem}

According to \Df~\ref{def:sensitivity}, the sensitivity of $G_{\theta \in
\Theta}$ is measured by its JFN. \Thm~\ref{thm:control} states that by pruning
points having the highest sensitivity in manifolds, we can increase the \vall\
of data produced from \gen. Moreover, this should \textit{not} notably undermine
\divv, given $p$, denoting percentage of out-of-manifold data, and
$\tau$, denoting percentage of samples with the highest sensitivity, are
\textit{close}. As a result, $\beta^\approx \approx \beta$. In particular, when
$\alpha$ is high, $\tau$ becomes negligible ($\tau$ is less than 0.01 in our setup;
see \S~\ref{sec:implementation} and \S~\ref{sec:evaluation}). This leads to a
tight lower bound (close to $\beta$) for $\beta^{\approx}$. We present
the full proof of \Thm~\ref{thm:control} at the end of this section.

\parh{The Mechanical Procedure.}~As will be shown in
\S~\ref{subsec:schemtic-div}, \divv\ can be largely increased by extensively
exploring the manifold and performing \gen. However, according to ``The None
Land'' theory, the upper bound of the \vall\ score $\alpha$ is inevitably
reduced (though not tight) when \divv\ increases. Therefore, to retain \vall, we
prune a proportion of $\tau$ samples having the highest sensitivity from the
outputs of \gen\ following \Thm~\ref{thm:control}.

\parh{Reflection on Previous \gen\ Setups.}~Key findings at this step can be
used to rigorously explain previous \gen\ setups. For instance, generative
model-based \gen\ would require to truncate the input space of the generator to
improve the realism (\vall) of generated samples. For the non-parametric input
mutation methods (i.e., $G_{\emptyset}$), the sensitivity is only decided by
$H_{\Delta} = H_{\max} - H_{\min}$. Thus, to improve \vall, we can shrink
the choices of $h$.\footnote{Note that to enable diversified mutation patterns, $H_{\min}$
is generally set to ``$- H_{\max}$'' in their setting.} Note that this is aligned with
their current setups, e.g., limiting the number of the mutated pixels, as
introduced in \S~\ref{subsec:motivation-validation}.

\begin{proof}[Proof of Theorem~\ref{thm:control}]
    For the data samples with the highest sensitivity $x \in S_1$,
    the proportion of out-of-manifold data is $p$, where
    $p = \frac{|S_2|}{|S|}$ and $\forall x \in S_2$, $x \in S_1$ and
    $x \notin \mathcal{M}$. The new \vall\ score $\alpha^\uparrow$ is
    calculated as below:
    \begin{equation*}
    \begin{aligned}
    \alpha^\uparrow &= \frac{\alpha|S| - (|S_1| - |S_2|)}{|S| - |S_1|}
    = \frac{\alpha|S| - \tau |S| + p |S|}{|S| - \tau |S|} \\
    &= \frac{\alpha - \tau + p}{1 - \tau}
    \end{aligned}
    \end{equation*}
    Similarly,
    \begin{equation*}
    \begin{aligned}
    \beta^\approx &= \frac{\alpha|S| - (|S_1| - |S_2|)}
    {\alpha |S| / \beta} = \frac{\alpha |S| - \tau |S| + p |S|}
    {\alpha |S|} \beta \\
    &= \frac{\alpha - \tau + p}{\alpha} \beta,
    \end{aligned}
    \end{equation*}
    according to \Lem~\ref{lem:oom}, $p > \tau(1 - \alpha)$,
    we thus have
    $$\alpha^\uparrow > \alpha, \qquad
    \beta^\approx > (1 - \tau) \beta$$
\end{proof}

\subsection{Increasing \divv: A Schematic Comparison}
\label{subsec:schemtic-div}

Our \gen\ is implemented as a practical framework, namely \tool, whose full
technical details are presented in \S~\ref{sec:design}. \tool\ mutates media
data by walking on the corresponding manifold towards different directions and
with varied step sizes. This enables diverse perceptual-level mutations and
gradually increases \divv.
\F~\ref{fig:manifold} presents a schematic comparison of \tool\ and existing
works. On a data manifold, a seed image $i$ is mutated when \tool\ walks toward
the bottom left and the top right. The footsteps of \tool, after being mapped
back to the high-dimensional space, yield a succession of mutated images with
perceptions changed.

Prior mutation-based works, as reviewed in \S~\ref{subsec:motivation-mutation},
directly mutate media data. Their mutated images typically surround the seed $i$
in an $\varepsilon$-radius sphere. Some mutated images breach the perceptual
constraints (e.g., by adding too much random noise), thus residing outside the
manifold (shown in \textcolor{pptred}{red} in \F~\ref{fig:manifold}; the
\textcolor{pptred}{red} dot is on the sphere and \textit{above} the manifold).
Although these mutated images may trigger DNN mis-predictions, they typically
stress only a limited range of behaviors of DNNs, given that perception changes
are not comprehensively explored.
Worse, we emphasize that the $\varepsilon$-radius must be \textit{reasonably
small}, as images may become unidentifiable, with disrupted senses, as
$\varepsilon$ rises. This conclusion can also be derived from
\Thm~\ref{thm:control}. \S~\ref{subsec:motivation-validation} described how
existing methods retain ``validity''; e.g., DeepHunter allows only one affine
transformation per image. Thus, existing methods are confined within limited
perceptions by design. Clearly, \tool\ can mimic existing mutations by limiting
the step size as smaller than $\varepsilon$.

\F~\ref{fig:manifold} also presents SOTA generative model-based methods, which
map images belonging to different manifolds to a unified space. As illustrated
in the ``The None Land'' theory, their methods are proved to only generate images
of negligible \vall\ for real-world media data.

\section{Design of \tool}
\label{sec:design}

\begin{figure*}[!ht]
  \centering
  \includegraphics[width=0.92\linewidth]{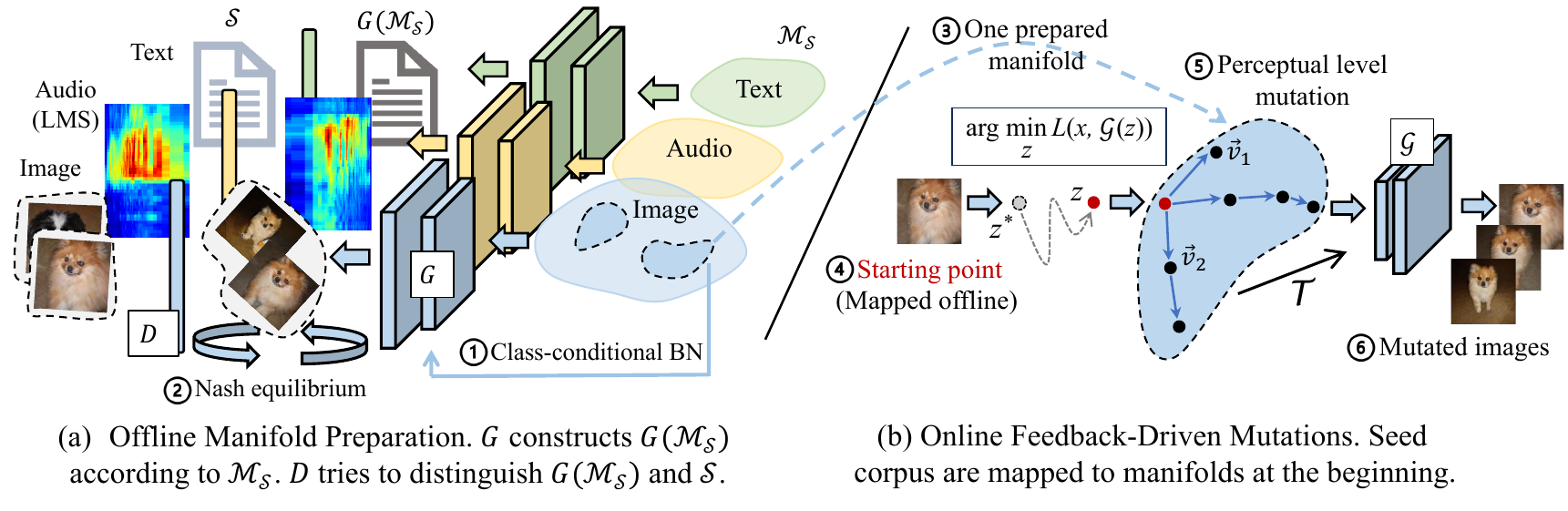}
  \vspace{-5pt}
  \caption{Workflow of \tool, which is unified for image, audio, and text.
  In the offline phase, \tool\ first constructs manifolds using GAN. It leverages
  class-conditional batch normalization to handle the separation among different
  manifolds. This process is finished when the generator $G$ and discriminator
  $D$ establish a Nash equilibrium. In the online phase, \tool\ performs perceptual
  mutation in each manifold with a footprint-aware manner. Guided by the testing
  objectives, \tool\ mutates certain perceptions that likely trigger DNN faults.}
%  \vspace{-5pt}
  \label{fig:workflow}
\end{figure*}

\F~\ref{fig:workflow} illustrates the high-level workflow of \tool, in terms of
manifold construction (\F~\hyperref[fig:workflow]{8(a)}) and feedback-driven
testing (\F~\hyperref[fig:workflow]{8(b)}). \S~\ref{subsec:gan} and
\S~\ref{subsec:mutation} will introduce these two phases respectively.

\mr{\parh{Application Scope and Position.}~\tool\ is designed as a DNN testing
tool that mainly addresses issues in the input mutation/generation. More
specifically, \tool\ seeks to improve the diversity of test inputs
(which are achieved via perceptual-level mutations) given a limited number of
seed inputs, thereby effectively and comprehensively studying DNN faults.
Meanwhile, \tool\ also aims to retain the validity of test inputs when performing
mutations, such that it discloses real DNN faults rather than those induced by
invalid inputs (i.e., false positives).

Techniques in \tool\ are based on data manifold, which is approximated using
generative models.
Overall, \tool\ demonstrates the successful application of data manifold in DNN
testing and does \textit{not} focus on proposing new methods for forming
manifold. \tool\ provides an out-of-box solution to leveraging generative models
to get (disconnected) manifolds for real-world media data with different
formats, in a unified way. The core techniques of \tool\ are agnostic to the
implementation of generative models or the format of media data.

We leverage GANs~\cite{goodfellow2014generative} in the current
implementation of \tool\ and discuss the design consideration in
\S~\ref{subsec:manifold-testing} (see \textbf{Generative Models}). Given that
said, we clarify that \tool\ is not specifically designed for GANs. Users may
replace our employed GAN with more advanced generative models, if needed. This
should be technically feasible: users only need to replace the mapping function
between media data and manifolds with that in the new generative models.}

\subsection{\mr{Manifold in DNN Testing}}
\label{subsec:manifold-testing}

A manifold can be constructed by finding
a projection $f_{\theta}: \mathcal{R}\footnote{
  Here we use $\mathcal{R}$ to for media data of different formats,
  including image, text, and audio.
} \rightarrow \mathcal{M}$ that converts
media data $x \in \mathcal{R}$ into $z=f_{\theta}(x)$ on $\mathcal{M}$. That said,
mapping $f_{\theta}(x) \in \mathcal{M}$ back to $x \in \mathcal{R}$ requires the
reconstructed data to be realistic. Also, $f_{\theta}: \mathcal{R} \rightarrow \mathcal{M}$
is often approximated on a limited number of data. This raises practical challenges
to construct an expressive manifold that can facilitate synthesizing diverse
$x \in \mathcal{R}$. \mr{It is shown that employing generative models can construct
manifolds~\cite{zhu2018image,holden2015learning,martinez2019studying}.
In general, these generative models are parameterized and enable ``learning''
data manifold from usually limited data samples. This makes generative models a
desirable choice for DNN testing where the seed inputs are typically limited.}

\mr{\noindent \textbf{Generative Models.}~Various paradigms of generative models
are proposed in past few years. For examples, variational auto-encoder
(VAE)~\cite{kingma2014auto} jointly trains an encoder and a decoder to form a
bidirectional mapping between media data and manifold. Nevertheless, outputs of
VAE suffer from the over-smooth issue~\cite{saito2017statistical}. Image details
such as edges and textures typically vanish during the generation procedure,
which will harm the effectiveness of DNN testing.
Flow-based~\cite{kingma2018glow} and Diffusion~\cite{ho2020denoising} models
adopt (mathematically) invertible modules to build the generative model. For
flow-based models, image quality remains a major concern, and they do not have a
low-dimensional latent space (the dimensions of latent space equal the image
size, i.e., $\#channels \times width \times height$). Also, despite diffusion
models can generate vivid and realistic images, their training requires billions
of training data. For instance, Stable Diffusion~\cite{diffuison} is trained on
5 billion images. We note that prior testing tools in this field and
\tool\ share a common design focus to effectively use (usually limited)
available seed inputs. Thus, employing Diffusion model indeed introduces a
dilemma for this setting: once developers have enough training data (i.e., the
seed inputs), they may no longer need to train the Diffusion model, as they can
directly use these billions of data (which are more than sufficient) to test the
target DNN. Moreover, collecting such a huge number of training data may be
likely infeasible for testing DNNs in various real-world, specialized domains,
e.g., medical image diagnosis or autonomous driving.

Therefore, \tool\ chooses generative adversarial nets (GANs)~\cite{goodfellow2014generative}
to form data manifold. Compared with VAE, GANs can generate images of higher quality.
It also requires much less training data than Diffusion models. A potential
limitation is that a GAN only contains one generator (decoder) that has maps
$\mathcal{M} \rightarrow \mathcal{R}$; it does not provide an encoder for
$\mathcal{R} \rightarrow \mathcal{M}$. 
In \S~\ref{subsec:gan}, we introduce how we build encoders for GANs.
}

\noindent \textbf{Requirement for Dataset.}~\F~\hyperref[fig:workflow]{8(a)} shows
the offline phase of \tool\ which prepares a manifold over dataset \ts. \tool\
mutates media data of different types. We use a GAN to create a specific
manifold over \ts, which can be images, audio, or text. We assume that \ts\
represents all accessible data for DNNs testing. In our evaluations (see
\S~\ref{sec:evaluation}), we use training data of the tested DNN as \ts\
to simulate testing conducted by the DNN developer. For general real-world
users of \tool, they may use any public/private dataset they have as \ts.

\noindent \textbf{Data Representation.}~Media data can be categorized into two
types: continuous and discrete. Image and audio data are continuous since they
are composed of floating numbers. Images are formed as $\#channels \times width
\times height$ matrices and their manifolds can be constructed directly on this
representation (see below). It is, however, challenging to capture perceptual
constraints of audio in the raw format (e.g., \texttt{.wav}). Generally, audio
can be converted into 2D forms, representing the acoustic information as
frequency series along the time and further transformed back to the
original ones losslessly. That is, an audio can be represented as an ``image''
of size $1 \times frequency \times time$. Our evaluation shows that the
\textit{image-style} representation is expressive, given that audio contents and
intonations can be characterized via frequencies, whose changes alter the formed
images in a constrained way. Thus, we convert audio into the log-amplitude of
Mel spectrum (LMS; see \F~\hyperref[fig:workflow]{8(a)}) and construct their
manifolds as for images.
Textual data comprise a sequence of words. They are discrete because there is no
``intermediate word.'' As noted in \S~\ref{subsec:media}, the word dependencies
encode the semantics and grammatical coherence of a sentence, which are the
primary focuses of their derived manifolds. Nevertheless, it is infeasible to
directly compute the manifold of text using GAN. We use
ARAE~\cite{zhao2018adversarially}, a prominent approach that uses adversarial
regularization, to convert discrete text into a continuous representation, enabling to
process text data with a GAN.

\subsection{Offline Manifold Preparation}
\label{subsec:gan}

\noindent \textbf{Preparing Manifolds via GANs.}~GAN employs generator network
$G$ and discriminator network $D$, with the objective to make data approximated
by $G$ indistinguishable under the view of $D$. GAN minimizes the ``distance''
between real and synthesized data distributions. It seeks a Nash equilibrium in
a two-player zero-sum game:

%\vspace{-10pt}
$$\min\limits_{G}\max\limits_{D}\mathbb{E}_{x \sim \mathcal{R}}[\log D(x)] +
\mathbb{E}_{z \sim p(z)} [\log(1 - D(G(z)))]$$
%\vspace{-10pt}

\noindent where $z$ is a vector of much lower dimensions and is sampled
from a continuous distribution $p(z)$, which is typically set as $\mathcal{N}(0,
I)$. $G$ implicitly encodes constraints formed by real data.
\tool\ first launches an offline phase to train GAN, including
generator network $G$ and discriminator network $D$, over a set of media data.
Only $G$ is used in the follow-up online testing. The training is finished until
$G$ can constantly generate diverse and valid media data instances.

\noindent \textbf{Handling Distinct Manifolds.}~Both $G$ and $D$ are composed of stacked
blocks~\cite{radford2015dcgan}. Each block has a convolutional layer, a batch
normalization (BN) layer~\cite{ioffe2015batch} and a non-linear function. The BN
layer acts as a regularizer by normalizing a batch of input for subsequent layers
as

%\vspace{-10pt}
$$x = \frac{x - Mean}{\sqrt{Var}}\gamma + \mu$$
%\vspace{-5pt}

\noindent with learned gain $\gamma$ and bias $\mu$, where $Mean$ and $Var$ are
the mean and variance of all $x \in \mathcal{R}$, respectively. It alleviates the problem
of \textit{internal covariate shift}, which refers to the varied distribution of
input for each layer during training, thus guaranteeing $G$ and $D$ to be of
saturating nonlinearity~\cite{ioffe2015batch}.

Data of different classes lie in distinct manifolds (e.g., dog vs.~vehicle
images); therefore, it is generally infeasible to design a universal model to
construct distinct manifolds, as proved in \Lem~\ref{lem:none}. Without training
a separate model for each manifold (which is computation-intensive), we use
class-conditional BN to alleviate this hurdle~\cite{de2017cbatch,brock2018biggan}.
The gain and bias in each BN layer depend on the manifold to be constructed.
\mr{And accordingly, the input to $G$ is the concatenation of a latent vector
$z$ and a class label\footnote{\mr{To ease reading, we omit the class label
when presenting the generator's input in the rest of this paper.}} of the target manifold.}
As illustrated in \F~\hyperref[fig:workflow]{8(a)}, we only need to switch between
BN layers to transit across distinct manifolds. Aligned with \S~\ref{sec:formal},
we denote the $G$, which considers the separation among manifolds via
class-conditional BN, as $\mathcal{G}$ in the rest of this paper. Accordingly,
the corresponding encoder is dubbed as $\mathcal{E} = \mathcal{G}^{-1}$. 

\mr{\noindent \textbf{Per-Input GAN Inversion for $\mathcal{R} \rightarrow \mathcal{M}$.}
A potential solution to getting $\mathcal{E} = \mathcal{G}^{-1}$ in GANs is building a
universal encoder (which can
map all data to the manifold) for a GAN's generator. Nevertheless, this strategy
is costly (since it almost doubles the original training cost) and also unnecessary.
Intuitively, before online testing, we only need to map the seed inputs onto
their coordinates in the manifold. Thus, to reduce the cost, we propose to implement
GAN inversion in a per-input granularity. Instead of building an encoder (as in
VAE), we re-form the inversion for an input (or a batch of inputs) as an optimization
task which can be solved using DNN training optimizers such as SGD~\cite{ruder2016overview}
or Adam~\cite{kingma2014adam}.

For a seed input $x$, GAN inversion aims to find its coordinate
$z \sim p(z)$ in the manifold such that $L(x, \mathcal{G}(z)) < \epsilon$, where
$L$ is the distance metric. This procedure can be formulated as the
following:

$$\mathcal{E}: \; \mathcal{G}^{-1}(x) = \arg\min_{z} L(x, \mathcal{G}(z))$$

\noindent To implement this, we start with a randomly selected $z^{*}$. The
optimizer then updates $z^*$ to $z$ to minimize $L(x, \mathcal{G}(z))$. The overall
procedure is similar to DNN training, but the optimizer is applied on $z$
(not DNN parameters). Since the latent space is much smaller than
the parameter space of a DNN, inverting GAN in a per-input granularity yields
significantly less cost than training an encoder.

Once $z = \mathcal{G}^{-1}(x)$ is obtained, by assigning a walking direction and a step size,
\tool\ walks from $z$ to $z'$. $z'$ will be fed to $\mathcal{G}$, a decoder $f_{\theta}^{-1}:
\mathcal{M} \rightarrow \mathcal{R}$, that maps $z'$ back to realistic and valid
high-dimensional media data $\mathcal{G}(z')$. Different walking directions will lead to
different perception changes and the step size decides the changing extent.
Since $p(z)$ is continuous, \tool\ can walk with an infinitely small step size.
}

\subsection{Online Feedback-Driven Mutation}
\label{subsec:mutation}

\F~\hyperref[fig:workflow]{8(b)} illustrates the online phase of \tool, with the
key algorithm in \A~\ref{alg:fuzz}. The offline phase has mapped a seed corpus
$\mathcal{S}$ to coordinates $\mathcal{C}$ (using $\mathcal{E}$ which considers
the separation among different manifolds) on manifold. A priority queue $\mathcal{Q}$
is initialized using all elements with equal priority in $\mathcal{C}$; each seed
$c$ in $\mathcal{Q}$ may be used as the starting point for several walking
iterations, unless its priority is lower than a threshold. The reason for
keeping each seed coordinate $c$ for multiple iterations is that mutating
different perceptions on the same data (e.g., images) is likely to trigger
different DNN behaviors.

Also, not all mutations can trigger new DNN behaviors --- mutating $c$ many
times is unlikely to reveal new faults. Therefore, we lower the priority of $c$
progressively (line 12) and spend more efforts on other seeds in $\mathcal{Q}$.

\newcommand\mycommfont[1]{\small\textcolor{pptgreen1}{#1}}
\SetCommentSty{mycommfont}
\begin{algorithm}[t]
\label{alg:fuzz}
\caption{Footprint-Aware Mutation.}
    Fuzzing Seed: $\mathcal{C}$; \tcp{coordinates on $\mathcal{M}$; mapped via $\mathcal{E}$}
    Priority Queue: $\mathcal{Q}$; \\
    Generative Model: $\mathcal{G}$; \\
    $\mathcal{Q} \gets \mathcal{C}$; \\
    $\lambda \gets 0, t \gets 1$; \\
    $\mu_0 \gets 0, \mathcal{T}_0 \gets 0$; \\
    $\eta_{m} \gets \max_{c \in \mathcal{C}} J(c)$; \tcp{record maximal JFN}
    
    \While{\textbf{not} can\_terminate()}{
    
      \uIf{uniform(0, 1) $\leq \lambda$}{
        $c \gets \mathcal{Q}.select()$; \\
        $c_t \gets c\mathcal{T}_{t-1}$; \tcp{footprint-aware mutation}
        $\mathcal{Q}.decrease\_priority(c)$; \\
      }
     \Else{$c_t \gets sample\_from\_manifold()$}
      \If{$J(c_t) < \eta_{m}$ \textbf{and} objective($\mathcal{G}(c_t)$)}{
        $\mu_t \gets \frac{t-1}{t}\mu_{t-1} + \frac{c_t}{t} $; \\
        % $\mathcal{T}_t \gets \frac{(t-1)\mathcal{T}_{t-1}}{t}$\;
        $\mathcal{T}_t \gets \frac{(t-1)\mathcal{T}_{t-1}}{t} + \frac{(t-1)(\mu_{t-1}-c_t)(\mu_{t-1}-c_t)^T}{t^2}$; \\
        $t \gets t + 1$; \\
        $\mathcal{Q}.add(c_t)$; \\
        $\lambda \gets \min(\lambda+\delta, \Lambda)$; \tcp{exploit vs. explore}
      }
    }  
\end{algorithm}

\noindent \textbf{Retaining \vall.}~Before explaining the online phase presented
in \F~\hyperref[fig:workflow]{8(b)}, we first discuss how to retain \vall\ in
online testing. As clarified in \S~\ref{subsec:tradeoff}, extensive mutations,
though increasing \divv, can undermine \vall\ of produced samples. Accordingly,
\S~\ref{subsec:tradeoff} presents a mechanical approach to retaining \vall\ by
discarding samples with high local sensitivity on manifolds. To do so, once
samples in the seed corpus are mapped to their manifolds, we record the maximal
JFN, dubbed $\eta_{m}$, of all seeds (line 7 in \A~\ref{alg:fuzz}; recall that
JFN characterizes the local sensitivity of seeds).
\mr{Then, during mutations, a mutated seed (or a randomly sampled coordinate;
see line 14) will be kept for further mutation only if its JFN is no larger than
$\eta_{m}$ (line 16).}

\noindent \textbf{Exploitation vs.~Exploration.}~Our mutation is inspired by
``exploitation vs.~exploration''~\cite{watkins1992q}, a popular paradigm that
enables random exploration (to uncover new knowledge) and exploits knowledge
learned from previous footprints.
At time step $t$, we may (with probability $\lambda$) mutate a seed $c$
with the highest priority in $\mathcal{Q}$ (lines 9--12). This way, \tool\
walks from $c$ to $c_t$ as $c_t = c\mathcal{T}$ where $\mathcal{T}$ denotes the
\textit{direction}, and its cardinality is the \textit{step size}.
\mr{$\mathcal{T}$ is a covariance matrix calculated over previous footprints
taken on a manifold during the current online phase, representing the learned knowledge.}
\tool\ may also perform random exploration by randomly walk to $c_{t}$ sampled
on the manifold with a probability $1 - \lambda$ (line 14).

Random exploration may trigger new findings that challenge DNN predictions.
Then, through repetitive explorations, $\mathcal{T}$ gradually converges
to directions that reflect effective perception changes (revealing new DNN
behaviors). Thus, we expect that mutations according to $\mathcal{T}$ may induce
more oriented explorations of perception changes on the manifold and trigger
more DNN behaviors (thus more likely to reveal DNN faults) in a given time budget.
Consider the \colorbox{pptblue}{manifold} in \F~\hyperref[fig:workflow]{8(b)}. If
the testing objective (line 16) is neuron coverage, suppose that
$c$ has been mutated twice using $\vec{v}_{1}$ and $\vec{v}_{2}$, which induces
the activation of distinct groups of neurons in the target DNN, namely $g_{1}$
and $g_{2}$. As \tool\ maintains $\mathcal{T}$, subsequent mutations over $c$
will use $\vec{v}_{1} + \vec{v}_{2}$, and therefore, mutations directing to
$\vec{v}_{1} + \vec{v}_{2}$ probably stress neurons interleaving with $g_{1}$
and $g_{2}$.

\noindent \textbf{\mr{Controlling the Trade-off.}} The trade-off between exploitation
vs. exploration is controlled using $\delta$ and $\Lambda$ (line 21). Starting from 0,
$\lambda$ is increased by $\delta$ each step if the testing objective is satisfied (line 16),
e.g., coverage increase,
until it reaches the predefined threshold $\Lambda$. This will gradually reduce the
effect of randomness in determining walking directions and step sizes.
\mr{With a large $\delta$, the online mutation will quickly converge to
exploitation but may not learn useful knowledge from a few explorations.
However, if $\delta$ is too small, the online phase will be dominated by
random explorations. Although it may comprehensively explore DNN behaviors,
such knowledge cannot be timely exploited for oriented mutations.
In addition, $\lambda$ is not allowed to exceed $\Lambda$ ($\Lambda < 1$) to
retain chances of finding new useful directions and step sizes on the manifold.
$1 - \Lambda$ decides the minimal ratios of random exploration in the online
phase; $\Lambda$ should be set as a relative high value to avoid harming the
exploitation. Meanwhile, we do not expect a too large $\Lambda$ as it will
eliminate exploring new DNN behaviors. See \S~\ref{sec:implementation} for value
suggestions of our employed hyperparameters.}

\noindent \textbf{Footprint-Aware Mutation.}~Given each coordinate $c_{i}$ on
the manifold as a vector, we keep track of all covered footsteps by using a
\textit{covariance matrix} $\mathcal{T}$. In lines 9--12, \tool\ mutates $c$
using $\mathcal{T}_{t-1}$, which records the footsteps taken on this manifold
until time $t-1$. $\mathcal{T}$, as a matrix, is initialized to 0. If the media
data $\mathcal{G}(c_{t})$ derived from step $c_{t}$ increases the DNN testing objective,
\tool\ adds $c_{t}$ to $\mathcal{Q}$ and updates $\mathcal{T}$ as follows:

%\vspace{-10pt}
$$\mathcal{T}_{t} = \frac{(t-1)\mathcal{T}_{t-1}}{t} + \frac{(t-1)(\mu_{t-1} - c_{t})(\mu_{t-1} - c_{t})^{\top}}{t^{2}}$$
%\vspace{-10pt}

\noindent where $\mu_{t-1} = \frac{1}{t-1}\sum^{t-1}_{i=1}c_{i}$ and will be
updated as $\mu_{t} = \frac{t-1}{t}\mu_{t-1} + \frac{c_{t}}{t}$.
The implementation of $\mathcal{T}$ is an incrementally updated (using each $c_{t}$)
covariance matrix. From a holistic view, $\mathcal{T}$ encodes the direction and the
corresponding step size on the manifold that are prone to satisfy DNN testing objective,
according to experience collected until time $t$.

$\mathcal{T}$ is initialized for each online phase. Effective walking directions
and step sizes (i.e., better enhancing DNN testing criteria) require a
$\mathcal{T}$ that is specific for a manifold and the target DNN. For instance,
to test an image classifier $m_1$ that relies on the ear shape to classify dogs,
it would be beneficial to mutate perceptions related to the ear within manifold
$\mathcal{M}_{dog}$ over dog images. In contrast, to test an object detector
$m_2$, mutating perceptions related to motions (e.g., standing vs.~lying) within
$\mathcal{M}_{dog}$ may better stress $m_2$. Furthermore,
\S~\ref{subsec:nn} has introduced the \textit{translation invariance}
property of a CNN. Thus, $\mathcal{T}$ helps to \textit{adaptively} recognize
certain perceptions that are prone to enhance DNN testing objective and reveal
new behaviors, and it saves efforts mutating certain perceptions that have been
shown as less effective based on current online phase experience.

\section{Implementation \& Evaluation Setup}
\label{sec:implementation}

\tool\ has about 2,100 LOC (counted by cloc~\cite{cloc}). \tool\ is written in
PyTorch 1.9.0. Both online and offline phases of \tool\ are launched on one
Intel Xeon CPU E5-2683 with 256GB RAM and one Nvidia GeForce RTX 2080 GPU.

% \smallskip
\noindent \textbf{\mr{Configurations.}}~For the current implementation of \tool, we set
$\delta$ and $\Lambda$ as 0.0005 and 0.8, as two parameters in our exploitation
(see \A~\ref{alg:fuzz}), respectively.
\mr{As explained in \S~\ref{subsec:mutation}, these two parameters
decide the trade-off between exploitation vs. exploration; we recommend
users setting $\delta$ to a small value (e.g., around $0.001$) and $\Lambda$
to a relatively large value (e.g., within $[0.75, 0.95]$).}
Also, as mentioned in \S~\ref{subsec:gan}, media
data of different sizes are mapped to manifold of different dimensions;
we introduce the details in each corresponding section in \S~\ref{sec:evaluation}.
For inverting $\mathcal{G}$ as in \S~\ref{subsec:gan}, we use Adam optimizer
with a learning rate $0.002$.

\parh{Pruning via Sensitivity.}~As in \A~\ref{alg:fuzz}, we prune a mutated input
if its JFN (see \Df~\ref{def:sensitivity}) is higher than the maximal JFN among
seed corpus. In our experiments, we find that less than $1\%$ mutated inputs are
pruned, i.e., $\tau < 0.01$ in \Thm~\ref{thm:control}. We interpret the findings
as encouraging: according to the discussion associated with
\Thm~\ref{thm:control}, \divv\ is preserved, and our mutations have high \vall.

\subsection{\mr{Baselines}}
\label{subsec:baseline}

% \smallskip
The target DNNs and the involved datasets are
introduced in each experiment in \S~\ref{sec:evaluation}. \tool\ is compared
with representative tools reviewed in \S~\ref{sec:motivation}. To generate
diverse and valid images at their best effort, all of these tools are configured
with some key hyper-parameters. We list their setups as follows:

\noindent \textbf{Generative Model-Based:}~\cite{kang2020sinvad}
and~\cite{dola2021distribution} convert all images into a unified latent space
and do \textit{not} consider the separation of real manifolds. As proved in
\S~\ref{subsec:validity}, they have negligible \vall\ for real images. We
configure and test them using two popular real-world image datasets, ImageNet
and CIFAR10. We report that they only generate images of meaningless color
blocks (see some examples in \F~\hyperref[fig:table]{5(b)}). As clarified
previously, no universal perceptual constraints apply to all
images~\cite{lee2007nonlinear,bengio2013representation}, e.g., what and how
perceptions constitute a car does not apply to portraits. We thus skip comparing
with them in the following experiments.

% \smallskip
\noindent \textbf{DeepHunter}~\cite{xie2018coverage} requires that \#mutated pixels
is less than $a \times$ \#total pixels, or the maximal value of changed
pixels is less than $b \times 255$. We follow its default setting where
$a=0.02$ and $b=0.2$.

% \smallskip
\noindent \textbf{TensorFuzz}~\cite{odena2018tensorfuzz} truncates the
accumulated noise within $L_{\infty}$. We use its shipped configuration which
sets $L_{\infty}$ to $0.4$.
    
% \smallskip
\noindent \textbf{DeepSmart}~\cite{demir2019deepsmartfuzzer} mutates one
small region of an image each time. We follow its default setting where
\#regions is nine. Since it does not explicitly preserve validity of mutated
images, we use the image validation scheme of DeepHunter.
    
% \smallskip
\noindent \textbf{DeepTest}~\cite{tian2018deeptest} rules out likely-broken
mutated images based on the mean squared error (MSE). Given an image $i$ and the
mutated output $i'$, we only keep $i'$ satisfying $MSE(i, i') < 1000$. Setting
MSE as $1000$ denotes a large and tolerable configuration with pixel values
in $[0, 255]$.

% \smallskip
\noindent \textbf{DeepRoad}~\cite{zhang2018deeproad}/\textbf{TACTIC}~\cite{li2021tactic}
is specifically designed for testing autonomous driving. We reuse its released model
to mutate driving scenes.

\subsection{\mr{Testing Objectives}}
\label{subsec:objectives}

% \smallskip
We benchmark \tool\ and the previous testing tools using different testing
objectives, as introduced below.

\noindent \textbf{\mr{Structural Coverage:}}~Discrete states over neuron outputs
are defined in structural neuron coverage. When being used as testing objectives,
they guide the testing to maximize the covered states (i.e., neuron coverage).
\mr{Compared with other testing objectives, structural neuron coverage can
\textit{reflect the testing comprehensiveness} as it offers interpretable values
within $[0, 100\%]$, which quantify the explored DNN behaviors.}
We consider the following representative structural coverage due to their high
computational efficiency\footnote{\mr{Their speed can be boosted via matrix computation;
we use the optimized implementations provided by~\cite{coverage}}.}.

% \smallskip
\noindent \underline{Neuron Coverage (NC)}~\cite{pei2017deepxplore} rescales
neuron outputs from the same layer to $[0, 1]$: a neuron is activated if its
scaled output is greater than a threshold $T$. Same
with~\cite{xie2018coverage,demir2019deepsmartfuzzer}, we set $T$ as $0.75$.
    
% \smallskip
\noindent \underline{K-Multisection Neuron Coverage
  (KMNC)}~\cite{ma2018deepgauge} decides the normal output range of each neuron
  using all training data. It then divides output range of each individual
  neuron into $K$-sections. The coverage is computed as how many (ratio) neuron
  outputs lie in the total \#neuron $\times~K$ sections. Following DeepHunter,
  we set $K$ as $1000$.
    
% \smallskip
\noindent \underline{Neuron Boundary Coverage (NBC)}~\cite{ma2018deepgauge}
calculates coverage by counting \#neuron whose output lies outside the range
decided in KMNC.
    
% \smallskip
\noindent \underline{Strong Neuron Activation Coverage
  (SNAC)}~\cite{ma2018deepgauge} deems a neuron as activated if its output
is greater than the upper bound of the output range decide in KMNC.
    
% \smallskip
\noindent \underline{Top-K Neuron Coverage (TKNC)}~\cite{ma2018deepgauge}
measures ratio of neurons in a layer appearing in the top-$K$ outputs.
We set $K$ to $10$.

\mr{
\noindent \textbf{Cluster-Based Coverage.}~The TensorFuzz paper also
proposes a cluster-based coverage metric for DNN testing, which is
often dubbed as TensorFuzz Coverage (TFC)~\cite{odena2018tensorfuzz}.
Different from structural coverage that focus on neurons and their states,
TFC treats all neuron outputs in a DNN as one high-dimensional vector $v$.
During testing, a new cluster is formed if the distances between a new
$v$ and the center of its nearest cluster is greater than $T$; the coverage
is counted as the number of clusters formed by a test suite. Unlike
structural coverage that set 100\% as the maximal coverage value, TFC does
not have a maximal coverage (\#clusters). In \S~\ref{sec:evaluation},
we set $T=5$ and $T=500$ for DNNs trained on CIFAR10 and ImageNet, respectively.
 
\noindent \textbf{Distribution-Aware Coverage.} NeuraL coverage
(NLC)~\cite{yuan2023revisiting} defines the coverage over distribution
formed by layer outputs of a DNN. Similar to TFC, NLC also does not set
a maximal coverage; it guides the testing to maximize the coverage value.
Despite being less interpretable than structural coverage, NLC manifests
better capability to reflect the diversity of test inputs and is more
effective to guide generating error-triggering inputs (see our results
in \S~\ref{sec:evaluation}). NLC does not have hyperparameter.

\noindent \textbf{Black-box Entropy.}~The above objectives are only
available for white-/gray-box testing. Thus, we also consider a black-box
testing objective to perform evaluation under a black-box setting, where
only DNN outputs can be observed during testing. This objective measures
the entropy between DNN outputs w.r.t. the original input and the mutated
inputs. It guides the testing to maximize the entropy. The black-box
entropy is also hyperparameter-free.
}

\begin{table}[t]
  \caption{\mr{Tested DNN models used in \S~\ref{sec:evaluation}.}}
  \label{tab:dnn}
  \vspace{-5pt}
  \centering
\resizebox{0.9\linewidth}{!}{
  \begin{tabular}{
    @{\hspace{1pt}}c@{\hspace{1pt}}|
    @{\hspace{1pt}}c@{\hspace{1pt}}|
    @{\hspace{1pt}}c@{\hspace{1pt}}|
    @{\hspace{1pt}}c@{\hspace{1pt}}
    }
    \hline
     Model         & \#Neuron & \#Layer  & Remark \\
    \hline
    ResNet50~\cite{he2016deep}     & 26570/27560*      & 54           & Non-sequential topology \\
    \hline
    VGG16~\cite{simonyan2014very}   & 12426/13416*      & 16           & Sequential topology \\ 
   \hline
    MobileNet-V2~\cite{howard2017mobilenets}  & 17066/18056*      & 53           & Mobile devices \\ 
    \hline
    Inception-V3~\cite{szegedy2016rethinking} & 10250/11240* & 121 & Feature extraction \\
    \hline
    DenseNet121~\cite{huang2017densely} & 17018/18216* & 95 & Extremely deep model \\
    \hline
    $\text{Dave}_{\text{orig}}$~\cite{pei2017deepxplore} & 1561 & 10 & Autonomous driving \\
    \hline
    $\text{Dave}_{\text{norm}}$~\cite{pei2017deepxplore} & 1561 & 10 & Autonomous driving \\
    \hline
    $\text{Dave}_{\text{dropout}}$~\cite{pei2017deepxplore} & 733 & 7 & Autonomous driving \\
    \hline
    Transformer$_\text{E2G}$~\cite{vaswani2017attention} & 168386 & 328 & Machine translation \\
    \hline
    Transformer$_\text{G2E}$~\cite{vaswani2017attention} & 168386 & 328 & Machine translation \\
    \hline
    Audio CNN~\cite{yuan2022automated} & 1610 & 7 & Audio classification \\
    \hline
    QMobileNet$^{**}_\text{Conv}$ & 18056 & 53 & Quantized Conv layer \\
    \hline
    QMobileNet$_\text{FC}$ & 18056 & 53 & Quantized Conv layer \\
    \hline
    QMobileNet$_\text{Conv+FC}$ & 18056 & 53 & Quantized Conv\&FC layer \\
    \hline
  \end{tabular}
  }
  \begin{tablenotes}
    \footnotesize
    \item *ImageNet version; each of the first five DNNs has two variants trained on
    ImageNet and CIFAR10.
    \item ** Three QMobileNet are black-box.
  \end{tablenotes}
 \vspace{-10pt}
\end{table}

\begin{table}[t]
  \caption{\mr{Datasets used in \S~\ref{sec:evaluation}.}}
 \vspace{-5pt}
  \label{tab:dataset}
  \centering
\resizebox{0.95\linewidth}{!}{
  \begin{tabular}{l|c|c|c|c}
    \hline
     Dataset   & Format & \#Class & \#Sample  & Size (Per-input) \\
    \hline
     CIFAR10~\cite{krizhevsky2009learning}   & Image & 10      & 50,000   & 3 $\times$ 32 $\times$ 32 pixels \\ 
    \hline
     ImageNet*~\cite{deng2009imagenet} & Image & 100    & 100,000 & 3 $\times$ 128 $\times$ 128 pixels\\ 
    \hline
    CityScape~\cite{cordts2016cityscapes}  & Image & N/A     & 2,759 & 3 $\times$ 256 $\times$ 128 pixels\\ 
    \hline
    SNLI~\cite{bowman2015large}      & Text  & N/A    & 570,000 & 10--20 words \\ 
    \hline
    SC09~\cite{donahue2019adversarial}       & Audio & 10    & 18,620 & $\sim$1000 ms \\ 
    \hline
  \end{tabular}
  }
  \begin{tablenotes}
    \footnotesize
    \item[*] *We randomly select 100 classes and rescale the size to
    128 $\times$ 128 for speedup.
  \end{tablenotes}
 \vspace{-10pt}
\end{table}

\subsection{DNNs and Datasets}
\label{subsec:dataset}

\mr{
\T~\ref{tab:dnn} lists the 19 DNNs tested in our evaluation.

\parh{General DNNs.}~The first ten DNNs (i.e., ResNet--DenseNet with each
of them trained on two datasets)
perform general image classification. They are representative in terms
of the structure and topology (e.g., depth, sequential vs. non-sequential),
featured platform (e.g., MobileNet is specifically designed for mobile
devices), and the common utility (e.g., Inception's outputs are commonly
employed to construct similarity metrics for images).

\parh{Specialized DNNs.}~The next six DNNs are widely adopted in
more specialized tasks for different formats of media datasets. The three
variants of Dave models predict steering angles for driving scenes
(i.e., a regression task). Transformer (which is the building block of
recent large language models) takes natural language sentences as inputs
and performs machine translation (e.g., translates English/German to
German/English).
Audio CNN is provided by~\cite{yuan2022automated} and recognizes the
contents of audios of human voices.

\parh{Black-Box DNNs.}~The last three DNNs are different variants of
quantized MobileNet. They are deployed in mobile devices and deemed
as black-box because only DNN outputs (i.e., the predicted probability
of all classes) are accessible. Testing these DNNs can demonstrate the
generalizability of \tool\ in black-box scenarios.

\parh{Datasets.}~\T~\ref{tab:dataset} lists the 5 datasets considered
in the evaluation. The first three are image datasets. In particular,
CIFAR10 and ImageNet are large-scale real-world datasets for general
image understanding. CityScape contains 2,759 images taken from the driver's view.
SC09 features 18,620 sound clips of human voices saying numbers from 0 to 9.
Each clip has a duration of about 1K ms and one number is said.
The Stanford Natural Language Inference (SNLI) dataset contains 570,000
human-written English sentences. SNLI is commonly used for natural language
inference and is grammatically coherent.
}

\section{Evaluation}
\label{sec:evaluation}

\mr{
We evaluate \tool\ by investigating the following research questions (RQs):
\textbf{RQ1 (Effectiveness):}~Is \tool\ more effective than previous tools?
\textbf{RQ2 (Quality):}~Can \tool\ generate test inputs of higher quality?
\textbf{RQ3 (Generalizability):}~Can \tool\ be generalized to different media
data, tasks, and working scenarios? These three RQs are studied in
\S~\ref{subsec:effectiveness}--\S~\ref{subsec:generalizability}, respectively.
}

\subsection{RQ1: Effectiveness}
\label{subsec:effectiveness}

\mr{
To answer \textbf{RQ1}, we evaluate the effectiveness of input mutation
and validation in \tool\ and previous tools. For different input mutation
strategies, we study if their generated inputs can comprehensively trigger
DNN behaviors and disclose more DNN faults. For input validation schemes,
we evaluate whether they can faithfully rule out invalid inputs. We also
benchmark the effectiveness of \tool's design considerations (i.e., the
footprint-aware mutation) via ablation studies.
}

\noindent \textbf{Settings.}~Following previous
works~\cite{xie2018coverage,demir2019deepsmartfuzzer,odena2018tensorfuzz}, we
focus on image classification. We use CIFAR10 and ImageNet in this section.
The dimensionality of each manifold is 120. At this step, training GAN takes
one day for ImageNet (due to large volume of data) and 2 hours for CIFAR10.
We release our trained models in~\cite{snapshot}.
All tested DNNs of CIFAR10 reach competitive test accuracy over 93\%.
ImageNet-trained DNNs are shipped by PyTorch.
\mr{We use all training data of CIFAR10/ImageNet as the seed corpus for mutation
or manifold construction. That is, all prior tools and \tool\ have the same
data source and \tool\ is \textit{not} empowered by more data. Overall,
prior tools and \tool\ are compared to see which one enables more effective
mutations by utilizing the same seeds.
}
We let all tools run $T$ hours and $T$ is set to $6$ and $4$ for DNNs trained
on ImageNet and CIFAR10, respectively. We set all tools to perform at most
$50$ mutations on an image in one iteration.

\mr{
\parh{Objectives \& Criteria.}~When being used as the testing objective, the
five structural coverage metrics (see \S~\ref{subsec:objectives}) are mainly adopted
for reflecting the testing comprehensiveness because they provide interpretable
numbers for the covered states of neurons. To study the disclosed DNN faults, 
we use NC (as one representative structural coverage\footnote{\mr{Other structural coverage
metrics, such as SNAC and NBC, can hardly guide generating error-triggering
inputs (prior tools have zero coverage increase under these metrics;
see \T~\ref{tab:coverage-rq1}). This issue was also pointed out in one
recent work~\cite{yuan2023revisiting}.}}), TFC, and NLC as the
testing objectives.
}

\begin{table}[t]
% \vspace{-10pt}
\caption{Ratios of invalid images without explicitly format restrictions.}
%\vspace{-10pt}
\label{tab:invalid-rq1}
\centering
\resizebox{0.92\linewidth}{!}{
\begin{tabular}{
  @{\hspace{2pt}}c@{\hspace{2pt}}|
  @{\hspace{2pt}}c@{\hspace{2pt}}|
  @{\hspace{2pt}}c@{\hspace{2pt}}|
  @{\hspace{2pt}}c@{\hspace{2pt}}|
  @{\hspace{2pt}}c@{\hspace{2pt}}|
  @{\hspace{2pt}}c@{\hspace{2pt}}
  }
    \hline
            & \tool\       & DeepHunter    & TensorFuzz    & DeepSmart & DeepTest  \\
    \hline
    Invalid & \textbf{0}   & 76.96\%       & 54.37\%       & 77.23\%   & 41.17\%    \\
    \hline 
\end{tabular}
}
%\vspace{-5pt}
\end{table}
  
\begin{table}[t]
  % \vspace{-10pt}
  \caption{Testing comprehensiveness (\%) before $\rightarrow$ \textcolor{pptred1}{after}
  excluding format-invalid images for ResNet trained on CIFAR10.}
  %\vspace{-10pt}
  \label{tab:format-com-rq1}
  \centering
  \resizebox{1.0\linewidth}{!}{
  \begin{tabular}{
    @{\hspace{2pt}}c@{\hspace{2pt}}|
    @{\hspace{2pt}}c@{\hspace{2pt}}|
    @{\hspace{2pt}}c@{\hspace{2pt}}|
    @{\hspace{2pt}}c@{\hspace{2pt}}|
    @{\hspace{2pt}}c@{\hspace{2pt}}|
    @{\hspace{2pt}}c@{\hspace{2pt}}
    }
    \hline
                  & NC & KMNC    & SNAC    & NBC & TKNC  \\
      \cline{2-6}
      Init. & 55.39 & 1.43 & 0 & 0 & 23.21 \\
      \hline \hline
      \tool\ & \textbf{64.37} $\rightarrow$ \textbf{64.37} & 2.98 $\rightarrow$ 2.98 & \textbf{3.27} $\rightarrow$ \textbf{3.27} & \textbf{5.99} $\rightarrow$ \textbf{5.99} & \textbf{26.24} $\rightarrow$ \textbf{26.24} \\
      \hline
      DeepHunter & 61.85 $\rightarrow$ \textcolor{pptred1}{60.62} & \textbf{7.29} $\rightarrow$ \textbf{\textcolor{pptred1}{3.99}} & 3.15 $\rightarrow$ \textcolor{pptred1}{0} & 3.15 $\rightarrow$ \textcolor{pptred1}{0} & 25.38 $\rightarrow$ \textcolor{pptred1}{24.43} \\
      \hline
      TensorFuzz & 56.55 $\rightarrow$ \textcolor{pptred1}{56.21} & 2.94 $\rightarrow$ \textcolor{pptred1}{1.52} & 3.23 $\rightarrow$ \textcolor{pptred1}{0} & 3.23 $\rightarrow$ \textcolor{pptred1}{0} & 23.51 $\rightarrow$ \textcolor{pptred1}{23.38} \\
      \hline
      DeepSmart & 60.85 $\rightarrow$ \textcolor{pptred1}{58.94} & 6.85 $\rightarrow$ \textcolor{pptred1}{2.74} & 3.07 $\rightarrow$ \textcolor{pptred1}{0} & 3.07 $\rightarrow$ \textcolor{pptred1}{0} & 24.92 $\rightarrow$ \textcolor{pptred1}{23.86} \\
      \hline
      DeepTest & 57.18 $\rightarrow$ \textcolor{pptred1}{56.89} & 2.98 $\rightarrow$ \textcolor{pptred1}{2.98} & 1.54 $\rightarrow$ \textcolor{pptred1}{0} & 1.54 $\rightarrow$ \textcolor{pptred1}{0} & 23.68 $\rightarrow$ \textcolor{pptred1}{23.55} \\
      \hline
  \end{tabular}
  }
  \begin{tablenotes}
    \footnotesize
    \item * ``Init.'' denotes coverage achieved by seeds. Results are averaged over 5
    runs with maximum standard deviation $<10^{-3}$ (i.e., $0.0x\%$).
  \end{tablenotes}
  %\vspace{-5pt}
\end{table}

\begin{table}[t]
  % \vspace{-10pt}
  \caption{Disclosed DNN faults (\#) before $\rightarrow$ \textcolor{pptred1}{after}
  excluding format-invalid images for ResNet trained on CIFAR10.}
  %\vspace{-10pt}
  \label{tab:format-fault-rq1}
  \centering
  \resizebox{0.7\linewidth}{!}{
  \begin{tabular}{
    @{\hspace{2pt}}c@{\hspace{2pt}}|
    @{\hspace{2pt}}c@{\hspace{2pt}}|
    @{\hspace{2pt}}c@{\hspace{2pt}}|
    @{\hspace{2pt}}c@{\hspace{2pt}}
    }
    \hline
       & NC &  NLC  & TFC    \\ 
      \hline \hline
      \tool\ & \textbf{4073} $\rightarrow$ \textbf{4073} & \textbf{21675} $\rightarrow$ \textbf{21675} & \textbf{1309} $\rightarrow$ \textbf{1309}  \\
      \hline
      DeepHunter & 3219 $\rightarrow$ \textcolor{pptred1}{2011} & 18371 $\rightarrow$ \textcolor{pptred1}{7446} & 947 $\rightarrow$ \textcolor{pptred1}{740} \\
      \hline
      TensorFuzz & 95 $\rightarrow$ \textcolor{pptred1}{40} & 2726 $\rightarrow$ \textcolor{pptred1}{1879} & 40 $\rightarrow$ \textcolor{pptred1}{24} \\
      \hline
      DeepSmart & 511 $\rightarrow$ \textcolor{pptred1}{283} & 19196 $\rightarrow$ \textcolor{pptred1}{10196} & 256 $\rightarrow$ \textcolor{pptred1}{120} \\
      \hline
      DeepTest & 1020 $\rightarrow$ \textcolor{pptred1}{697} & 2172 $\rightarrow$ \textcolor{pptred1}{1007} & 1.54 $\rightarrow$ \textcolor{pptred1}{411} \\
      \hline
  \end{tabular}
  }
  %\vspace{-5pt}
\end{table}

\subsubsection{Input Validation}
\label{subsubsec:eval-validation}

As introduced in
\S~\ref{subsec:motivation-validation}, existing tools use heuristics to ensure
the realism of mutated images. \mr{Nevertheless, we find that their heuristics may
overlook images violating legit formats of RGB images (i.e., pixels values must lie
in $[0, 255]$)}, let alone perception \vall.

\noindent \textbf{Format-Invalid Inputs.}~For each testing tool, we count how many
format-invalid images are overlooked by merely using its input validation scheme
(without explicit considering the format restrictions). \T~\ref{tab:invalid-rq1}
presents the ratios of format-invalid images generated by prior tools when mutating
images from ImageNet. The percentages are high for all existing tools, indicating
the ineffectiveness of their validation schemes and their mutations that directly
operate on image pixel bytes. \tool, in contrast, mutates perceptions of media data, %(within manifolds),
which by design would not break the format.

\noindent \textbf{Misleading Testing.}~These neglected format-invalid images can
be misleading as they can falsely manifest a high testing comprehensiveness
and also report some false DNN faults. To unveil this issue\footnote{\mr{Format-invalid
``images'' can still be saved as valid image files, because common image libraries
truncate the pixel values into $[0, 255]$ before saving.}}, we study how the testing
comprehensiveness and \#faults change after filtering out format-invalid images.
Results are given in \T~\ref{tab:format-com-rq1} and \T~\ref{tab:format-fault-rq1},
respectively. As noted, the testing comprehensiveness and \#faults is
reduced considerably in previous tools. \mr{While this issue can be easily fixed by
truncating the pixel values within $[0, 255]$ for RGB images, it is far more
challenging for media data of complex formats (e.g., audios) where domain
expertises are required. These observations highlight the superiority of
perceptual-level mutations implemented in \tool, given that it is unified and agnostic
to certain format restrictions. We further evaluate the perception validity
\vall\ in \S~\ref{subsec:fault}.}

\begin{table*}[t]
  % \vspace{-10pt}
  \caption{\mr{Testing comprehensiveness (after excluding format-invalid images) achieved by different tools.}}
%   \vspace{-10pt}
  \label{tab:coverage-rq1}
  \centering
\resizebox{1.0\linewidth}{!}{
  \begin{tabular}{l|c|c|c|c|c|c|c|c|c|c|c}
    \hline
    \multirow{2}{*}{Criteria} & \multirow{2}{*}{Tool} & \multicolumn{5}{c|}{ImageNet}                                   & \multicolumn{5}{c}{CIFAR10} \\
    \cline{3-12} 
                              &                       & ResNet50 & VGG16 & MobileNet-V2 & DenseNet121 & Inception-V3   & ResNet50 & VGG16 & MobileNet-V2 & DenseNet121 & Inception-V3 \\
     \hline \hline
     \multirow{6}{*}{NC}      & Init.      & 59.73\% & 86.27\% & 88.15\% & 77.24\% & 82.60\% & 55.39\% & 41.75\% & 60.61\% & 61.94\% & 50.60\% \\
     \hline         
                              & \tool\     & \textbf{66.91\%} & \textbf{88.75\%} & \textbf{90.05\%} & \textbf{82.37\%} & \textbf{87.39\%} & \textbf{64.37\%} & \textbf{50.71\%} & \textbf{64.01\%} & \textbf{68.72\%} & \textbf{54.39\%} \\   
                              & DeepHunter & 62.52\% & 82.45\% & 88.59\% & 78.61\% & 83.93\% & 60.62\% & 47.40\% & 61.26\% & 64.98\% & 51.37\% \\
                              & TensorFuzz & 60.10\% & 80.41\% & 88.21\% & 77.50\% & 83.12\% & 56.21\% & 42.58\% & 62.18\% & 62.16\% & 50.94\% \\
                              & DeepSmart  & 60.65\% & 77.04\% & 88.40\% & 77.76\% & 83.00\% & 58.94\% & 45.98\% & 61.17\% & 62.98\% & 51.29\% \\
                              & DeepTest   & 60.40\% & 76.39\% & 88.18\% & 77.31\% & 82.81\% & 56.89\% & 43.54\% & 60.69\% & 62.24\% & 50.77\% \\
    \hline \hline
     \multirow{6}{*}{KMNC}    & Init.      & 0.59\% & 0.33\% & 0.58\%          & 0.59\% & 0.59\% & 1.43 & 1.10 & 1.40 & 1.32 & 1.35 \\
     \hline         
                              & \tool\     & \textbf{3.04\%} & \textbf{2.92\%} & 3.11\%          & \textbf{3.05\%} & \textbf{2.88\%} & 2.98\%        & \textbf{2.83\%} & 2.82\%          & \textbf{2.95\%} & \textbf{2.70\%} \\   
                              & DeepHunter & 1.93\%          & 1.81\%          & \textbf{3.32\%} & 3.02\%          & 2.63\% & \textbf{3.99\%} & 2.50\% & \textbf{3.75\%} & 2.22\% & 2.21\% \\
                              & TensorFuzz & 1.56\%          & 1.75\%          & 2.01\%          & 1.99\%          & 1.97\% & 1.52\%          & 1.78\% & 1.79\%          & 1.44\% & 1.39\% \\
                              & DeepSmart  & 2.19\%          & 1.82\%          & 1.77\%          & 2.13\%          & 1.89\% & 2.74\%          & 2.62\% & 2.78\%          & 2.02\% & 1.63\% \\
                              & DeepTest   & 0.79\%          & 1.80\%          & 1.52\%          & 1.30\%          & 1.39\% & 2.98\%          & 2.58\% & 1.87\%          & 2.38\% & 2.36\% \\
    \hline \hline
     \multirow{6}{*}{SNAC}    & Init.      & 0 & 0 & 0 & 0 & 0 & 0 & 0 & 0 & 0 & 0 \\
     \hline         
                              & \tool\     & \textbf{2.63\%} & \textbf{5.13\%} & \textbf{3.18\%} & \textbf{3.92\%} & \textbf{7.72\%} & \textbf{3.27\%} & \textbf{7.44\%} & \textbf{2.38\%} & \textbf{3.31\%} & \textbf{5.12\%} \\   
                              & DeepHunter & 0 & 0 & 0 & 0 & 0 & 0 & 0 & 0 & 0 & 0 \\
                              & TensorFuzz & 0 & 0 & 0 & 0 & 0 & 0 & 0 & 0 & 0 & 0 \\
                              & DeepSmart  & 0 & 0 & 0 & 0 & 0 & 0 & 0 & 0 & 0 & 0 \\
                              & DeepTest   & 0 & 0 & 0 & 0 & 0 & 0 & 0 & 0 & 0 & 0 \\
    \hline \hline
     \multirow{6}{*}{NBC}     & Init.      & 0 & 0 & 0 & 0 & 0 & 0 & 0 & 0 & 0 & 0 \\
     \hline         
                              & \tool\     & \textbf{2.77\%} & \textbf{5.22\%} & \textbf{3.31\%} & \textbf{3.89\%} & \textbf{7.63\%} & \textbf{5.99\%} & \textbf{3.83\%} & \textbf{2.84\%} & \textbf{4.15\%} & \textbf{5.79\%} \\   
                              & DeepHunter & 0 & 0 & 0 & 0 & 0 & 0 & 0 & 0 & 0 & 0 \\
                              & TensorFuzz & 0 & 0 & 0 & 0 & 0 & 0 & 0 & 0 & 0 & 0 \\
                              & DeepSmart  & 0 & 0 & 0 & 0 & 0 & 0 & 0 & 0 & 0 & 0 \\
                              & DeepTest   & 0 & 0 & 0 & 0 & 0 & 0 & 0 & 0 & 0 & 0 \\
    \hline \hline
     \multirow{6}{*}{TKNC}    & Init.      & 52.25\% & 82.02\% & 70.12\% & 76.58\% & 87.23\% & 23.21\% & 23.24\% & 26.64\% & 57.46\% & 22.77\% \\
     \hline         
                              & \tool\     & \textbf{57.43\%} & \textbf{86.92\%} & \textbf{76.09\%} & \textbf{81.32\%} & \textbf{90.11\%} & \textbf{26.24\%} & \textbf{32.59\%} & \textbf{28.13\%} & \textbf{60.48\%} & \textbf{25.43\%} \\   
                              & DeepHunter & 53.92\% & 83.77\% & 71.97\% & 78.01\% & 87.88\% & 24.43\% & 28.07\% & 27.25\% & 57.87\% & 22.58\% \\
                              & TensorFuzz & 52.50\% & 82.43\% & 70.97\% & 77.10\% & 87.30\% & 23.38\% & 23.77\% & 26.77\% & 57.63\% & 22.88\% \\
                              & DeepSmart  & 53.06\% & 82.92\% & 71.56\% & 77.81\% & 87.50\% & 23.86\% & 26.38\% & 26.98\% & 57.61\% & 23.19\% \\
                              & DeepTest   & 52.47\% & 82.23\% & 70.83\% & 76.79\% & 87.25\% & 23.55\% & 24.94\% & 26.77\% & 57.57\% & 22.96\% \\   
    \hline
  \end{tabular}
  }
  \begin{tablenotes}
    \footnotesize
    \item * ``Init.'' denotes coverage achieved by seeds. Results are averaged over 5
    runs with maximum standard deviation $<10^{-3}$ (i.e., $0.0x\%$).
  \end{tablenotes}
  \vspace{-5pt}
\end{table*}

\begin{table*}[t]
  % \vspace{-10pt}
  \caption{\mr{Disclosed DNN faults (after excluding format-invalid images) using different tools.}}
%   \vspace{-10pt}
  \label{tab:fault-rq1}
  \centering
\resizebox{1.0\linewidth}{!}{
  \begin{tabular}{
    l|c|c|c|c|c|c|c|c|c|c|c
    % l|
    % @{\hspace{2pt}}c@{\hspace{2pt}}|
    % @{\hspace{2pt}}c@{\hspace{2pt}}|
    % @{\hspace{2pt}}c@{\hspace{2pt}}|
    % @{\hspace{2pt}}c@{\hspace{2pt}}|
    % @{\hspace{2pt}}c@{\hspace{2pt}}|
    % @{\hspace{2pt}}c@{\hspace{2pt}}|
    % @{\hspace{2pt}}c@{\hspace{2pt}}|
    % @{\hspace{2pt}}c@{\hspace{2pt}}|
    % @{\hspace{2pt}}c@{\hspace{2pt}}|
    % @{\hspace{2pt}}c@{\hspace{2pt}}|
    % @{\hspace{2pt}}c@{\hspace{2pt}}
    }
    \hline
    \multirow{2}{*}{Criteria} & \multirow{2}{*}{Tool} & \multicolumn{5}{c|}{ImageNet}                                   & \multicolumn{5}{c}{CIFAR10} \\
    \cline{3-12} 
                              &                       & ResNet-50 & VGG16 & MobileNet-V2 & DenseNet121 & Inception-V3   & ResNet-50 & VGG16 & MobileNet-V2 & DenseNet121 & Inception-V3 \\
     \hline \hline        
     \multirow{5}{*}{NC}      & \tool\     & \textbf{19679} & \textbf{9932} & \textbf{2591} & \textbf{3012} & \textbf{2287} & \textbf{4073} & \textbf{3044} & \textbf{799} & \textbf{1221} & \textbf{710} \\   
                              & DeepHunter & 7455 & 5139 & 2531 & 2201 & 1196 & 2011 & 1714 & 631 & 630 & 419 \\
                              & TensorFuzz & 783 & 116 & 93 & 17 & 23 & 40 & 12 & 8 & 0 & 2 \\
                              & DeepSmart  & 3963 & 2226 & 1118 & 939 & 1004 & 283 & 219 & 78 & 54 & 52 \\
                              & DeepTest   & 614 & 662 & 102 & 132 & 171 & 697 & 690 & 162 & 202 & 279 \\
    \hline \hline               
     \multirow{5}{*}{NLC}     & \tool\     & \textbf{65387} & \textbf{86361} & \textbf{86226} & \textbf{53955} & \textbf{82313} & \textbf{21675} & \textbf{62765} & \textbf{35209} & \textbf{17233} & \textbf{25524} \\   
                              & DeepHunter & 22311 & 11368 & 32691 & 11023 & 15692 & 7446 & 10043 & 9564 & 8789 & 9704 \\
                              & TensorFuzz & 4979 & 4899 & 3128 & 4376 & 2287 & 1879 & 2876 & 2560 & 2340 & 1524 \\
                              & DeepSmart  & 29336 & 10676 & 33391 & 12201 & 13132 & 10196 & 22463 & 9917 & 8053 & 9177 \\
                              & DeepTest   & 4001 & 3367 & 5022 & 4189 & 4115 & 1007 & 886 & 1389 & 1053 & 1036 \\
     \hline \hline 
     \multirow{5}{*}{TFC}     & \tool\     & \textbf{3209} & \textbf{7322} & \textbf{3063} & \textbf{3175} & \textbf{3330} & \textbf{1309} & \textbf{2128} & \textbf{882} & \textbf{771} & \textbf{862} \\   
                              & DeepHunter & 1733 & 2871 & 993 & 901 & 1122 & 740 & 1395 & 563 & 364 & 601 \\
                              & TensorFuzz & 96 & 17 & 21 & 11 & 51 & 24 & 7 & 17 & 7 & 32 \\
                              & DeepSmart  & 320 & 599 & 361 & 98 & 167 & 120 & 256 & 101 & 34 & 59 \\
                              & DeepTest   & 1199 & 1310 & 433 & 307 & 511 & 411 & 504 & 212 & 301 & 371 \\
    \hline
  \end{tabular}
  }
  \vspace{-5pt}
\end{table*}

\subsubsection{Input Mutation Strategies}

\noindent \textbf{Testing Comprehensiveness.}~\T~\ref{tab:coverage-rq1} reports
the increased (structural) coverage of different tools, which can reflect the
testing comprehensiveness achieved by different input mutation strategies.
As just noted in \S~\ref{subsubsec:eval-validation}, prior tools can yield
format-invalid images but pass their validation. DNNs do not feature ``format
checking.'' These format-invalid images, when being fed to DNNs, can falsely
increase coverage value.
To clarify, we exclude coverage due to invalid images and report the
recalculated values in \T~\ref{tab:coverage-rq1}.

\begin{figure}[!ht]
  \centering
  %\vspace{-10pt}
  %\hspace{5pt}
  \includegraphics[width=1.00\linewidth]{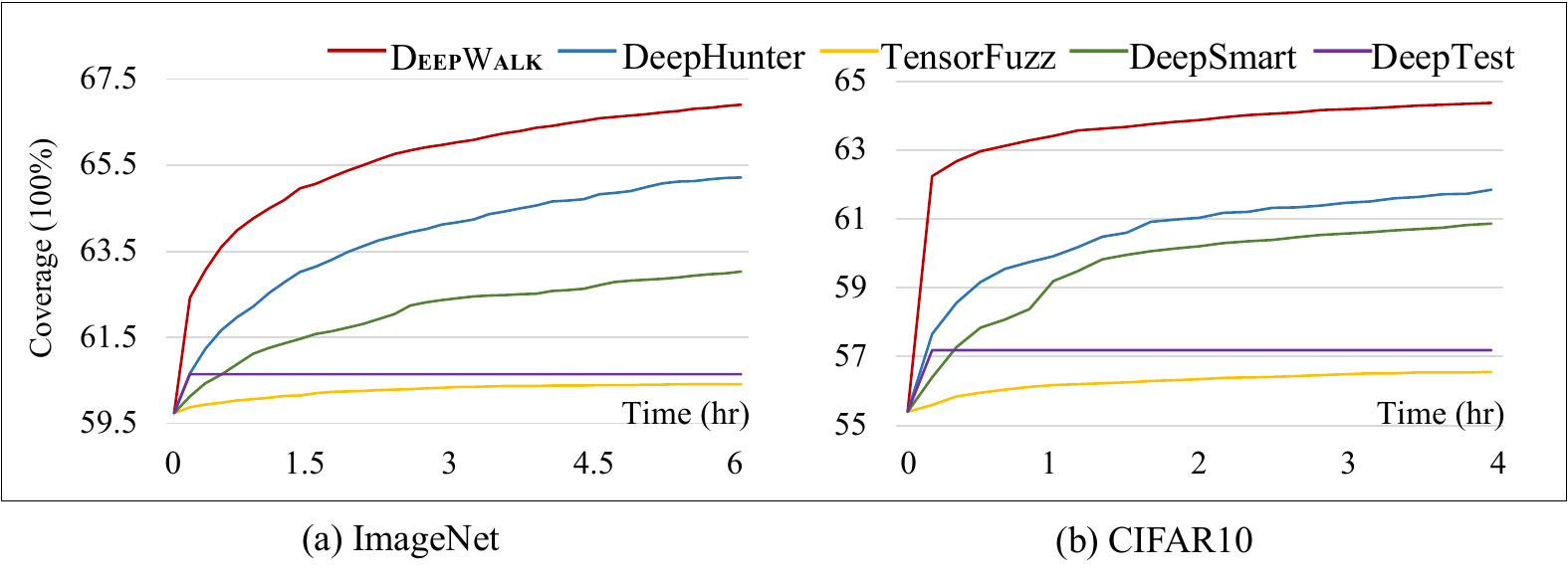}
 %\vspace{-5pt}
  \caption{Coverage (NC) increase of different tools on ResNet-50.}
  \label{fig:coverage-rq1}
  %  \vspace{-5pt}
\end{figure}

\tool\ outperforms earlier methods in most cases (even before excluding their
format-invalid outputs; see \T~\ref{tab:format-com-rq1}). This shows that
\tool's mutations are better for exploring DNNs. In \F~\ref{fig:coverage-rq1},
we show how the coverage value in ResNet50 increases with NC on ImageNet and
CIFAR10. \mr{\tool\ consistently achieves higher coverage during the testing process.
Also, the coverage values tend to be stable; previous tools are unlikely to surpass
\tool\ given more time budget.} DeepTest terminates earlier
than the other methods because it has exhausted all seeds: it discards a seed
even if the seed increases coverage.

\mr{\noindent \textbf{Interpretations.}~For KMNC in \T~\ref{tab:coverage-rq1},
DeepHunter performs better than \tool\ for three settings
(i.e., ImageNet \& MobileNet-V2, CIFAR10 \& ResNet50, and CIFAR10 \& MobileNet-V2).
Recall \F~\ref{fig:manifold} illustrates that existing tools explore an $\varepsilon$-radius
sphere surrounding a seed. KMNC, in general, divides one neuron's output range determined
by the initial seeds into $K$-sections. Therefore, by intensively exploring the
sphere near the seeds, DeepHunter can cover most sections checked
by KMNC. However, for criteria that reflect the ``corner-case'' behavior of test
inputs, such as SNAC and NBC, all prior tools have zero increase.
In contrast, \tool\ walks on the entire manifold (not just a sphere surrounding
a seed), and it performs much better for SNAC and NBC. TKNC has similar coverage
trending to that of NC because both focus on certain neurons with the highest
outputs.
}

% \smallskip
\mr{
\noindent \textbf{Low Coverage Increase.}~\tool\ exhibits relatively low increases on KMNC,
SNAC, and NBC. DNN outputs were close to zero (similar to a
normal distribution) due to input normalization and the presence of BN layers
(which convert neuron outputs to zero-centered small values). Such small and
normalized neuron outputs hardly reflect changes on these three fine-grained
coverage criteria, particularly for SNAC and NBC which look for edge-case
outputs.
Also, model training datasets (e.g., ImageNet) are comprehensive. Therefore,
most neuron output ranges are already covered when using the training data samples,
which  impedes testing tools to further increase coverage.
As in \T~\ref{tab:coverage-rq1}, this is more obvious for DNNs having higher
initial coverage (e.g., NC \& ImageNet \& MobileNet-V2).
}

\mr{
\noindent \textbf{Disclosed Faults.}~\T~\ref{tab:fault-rq1} lists
the number of disclosed DNN faults when using different testing objectives.
We only count an image as an error-triggering image if it is format valid.
As introduced in \S~\ref{sec:implementation}, we use TFC and NLC, two non-structural
coverage metrics, as the testing objectives when comparing the \#faults disclosed
by different tools. We also consider NC because it is one representative structural
coverage. For all three objectives, \tool\ triggers the most faults for all DNNs,
nearly doubling the amount of \#faults found by DeepHunter in the majority of cases.
When cross comparing DNNs trained on different datasets, we find that more faults
are triggered for the ImageNet cases. This is reasonable, given that the large scale of
ImageNet (1000 classes) makes the classification harder than that of CIFAR10 (10 classes).
Moreover, we also note that when guided by NLC, all tools (including \tool)
trigger more faults for all DNNs than the other two objectives. This result is consistent
with observations in recent research as NLC can more accurately respond to
DNN behaviors~\cite{yuan2023revisiting}.
}

\begin{table}[t]
  \caption{\mr{Ablation Results of \tool\ w/o $\mathcal{T}$.}}
  \label{tab:ablation}
%   \vspace{-10pt}
  \centering
\resizebox{1.0\linewidth}{!}{
  \begin{tabular}{c||c|c|c|c|c||c|c|c
    }
    \hline
    & \multicolumn{5}{c||}{Comprehensiveness (\%)} & \multicolumn{3}{c}{Faults (\#)} \\
    \hline
    \multirow{2}{*}{CIFAR10} & NC & KMNC & SNAC & NBC & TKNC & NC & NLC & TFC \\
                             & 60.94\% & 1.88\% & 0 & 0 & 24.50\% & 2265 & 11230 & 703  \\
    \hline
    \multirow{2}{*}{ImageNet} & NC & KMNC & SNAC & NBC & TKNC & NC & NLC & TFC \\
                             & 62.48\% & 1.95\% & 0 & 0 & 53.88\% & 7337 & 32134 & 1701 \\
    \hline
  \end{tabular}
  }
  % \vspace{-10pt}
\end{table}

\begin{figure*}[!ht]
  \centering
  \includegraphics[width=0.95\linewidth]{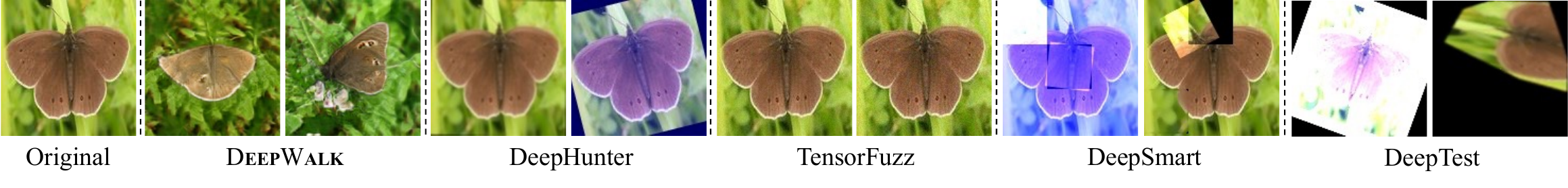}
 \vspace{-10pt}
  \caption{Qualitative evaluation of existing tools and \tool.}
% \vspace{-15pt}
  \label{fig:qualitative-rq1}
\end{figure*}

\subsubsection{Ablation Study}
\label{subsubsec:ablation}

To benchmark the ``exploitation vs.~exploration'' strategy adopted by \tool\
(\S~\ref{subsec:mutation}), we launch an ablation study to compare \tool\ with
random walking (i.e., \tool\ w/o $\mathcal{T}$) on ResNet50. \mr{In line with the
above evaluation, we conduct this ablation study across different coverage
criteria.} Results are presented in \T~\ref{tab:ablation}. Compared with
\T~\ref{tab:coverage-rq1} and \T~\ref{tab:fault-rq1}, walking on manifolds to
mutate perceptions, though randomly, is already comparable to DeepHunter (the
best in the competitors). This illustrates that perceptual-level mutations can
generally and effectively stress DNNs. Moreover, \tool\ (the ResNet50 columns in
\T~\ref{tab:coverage-rq1} and \T~\ref{tab:fault-rq1}) notably outperforms random
walking, showing the superiority of our walking strategy: it discovers useful
perceptions and the desirable extent of changes from historical footprints.

\subsection{Quality}
\label{subsec:fault}

\mr{
We answer \textbf{RQ2} in \S~\ref{subsec:fault}.
The quality of generated inputs is evaluated from the followings: \ding{172}
recognizability, i.e., mutated images retain original labels, \ding{173} 
diversity of the triggered faults (as faults of higher diversity are more
likely due to different root causes), and \ding{174} usefulness of
error-triggering inputs for repairing (i.e., model accuracy increases after
retraining with error-triggering inputs). From the empirical perspective, we note that
\ding{172} reflects the \vall\ of mutated images, and \ding{173} reflects the
\divv\ of mutated (error-triggering) inputs. \ding{174} reflects both \divv\ and
\vall: the reason is that to increase the accuracy of the model, we demand the
training data to be of high quality and diversity.
}

\subsubsection{Recognizability}
\label{subsubsec:recognizability}

\noindent \textbf{\mr{Qualitative Comparison.}}~\F~\ref{fig:qualitative-rq1} shows a
comparison of mutated images. As DeepHunter limits the number of affine
transformations applied on an image, its mutated images are similar to the
original one. TensorFuzz limits the accumulated noise on one image within an
$L_{\infty}$, retaining image perception. However, TensorFuzz generates outputs
that are less diversified because its mutation patterns are limited. DeepSmart
deconstructs images as it performs region-level mutations. DeepTest does not
explicitly confine mutation operations. As in \F~\ref{fig:qualitative-rq1}, it
can apply affine transformations repeatedly on one image, resulting in an
aberrant output. However, \tool\ mutates perceptions and produces diverse and
perceptually meaningful outputs. \mr{Note that images mutated by \tool\ have
relatively blurry backgrounds, as backgrounds are less associated with the
perceptual constraints in manifolds.} We present hundreds of \tool's outputs for
the reference at~\cite{snapshot}.

\begin{table}[t]
  %  \vspace{-5pt}
    \caption{Recognizability of mutated ImageNet images.}
    %\vspace{-10pt}
    \label{tab:input-rq1}
    \centering
  \resizebox{0.9\linewidth}{!}{
    \begin{tabular}{
      @{\hspace{2pt}}c@{\hspace{2pt}}|
      @{\hspace{2pt}}c@{\hspace{2pt}}|
      @{\hspace{2pt}}c@{\hspace{2pt}}|
      @{\hspace{2pt}}c@{\hspace{2pt}}|
      @{\hspace{2pt}}c@{\hspace{2pt}}|
      @{\hspace{2pt}}c@{\hspace{2pt}}
      }
      \hline
                    & \tool\       & DeepHunter    & TensorFuzz    & DeepSmart & DeepTest  \\
       \hline
       IS           & \textbf{33.87$\pm$0.45} & 21.21$\pm$0.48  & 21.24$\pm$0.94  & 23.15$\pm$0.42  & 10.46$\pm$1.36        \\
       \hline
       FID          & \textbf{54.31}         & 70.13         & 75.09          & 67.06             & 119.89        \\
      \hline
    \end{tabular}
    }
  % \vspace{-10pt}
\end{table}

\noindent \textbf{\mr{Quantitative Results.}}~It's challenging to manually exam the recognizability of mutated images given the large
volumes. Humans may also disagree with the recognizability (see the ``DeepTest'' cases
in \F~\ref{fig:qualitative-rq1}). Two popular criteria, inception score
(IS)~\cite{salimans2016improved} and Fr$\grave{e}$chet Inception Distance
(FID)~\cite{heusel2017gans}, are widely adopted in the AI community to assess
recognizability of images by comparing them with known collections of real
images.
FID calculates distances from a \textit{distribution} perspective and IS measures
the similarity via a \textit{class-aware} manner.
IS/FID are suitable for our setting since we aim to check whether mutated images
retrain their labels. In particular, we use them to measure how close the mutated
images are towards the seeds.
A smaller distance or higher similarity indicates that the mutated images are more
recognizable and presumably retain the labels.
The results are in \T~\ref{tab:input-rq1}. A higher IS is better, while a lower
FID suggests more recognizability. Though earlier techniques launch input validations to
keep realistic images, the images they regard as ``realistic'' manifest less
recognizability than images mutated by \tool. Furthermore, we see these results as proof
that, \tool, by addressing media data mutation and validation as a whole (via
manifold), is much better for both input mutation and validation.

\begin{figure}[!ht]
  \centering
  % \vspace{-5pt}
  \includegraphics[width=0.7\linewidth]{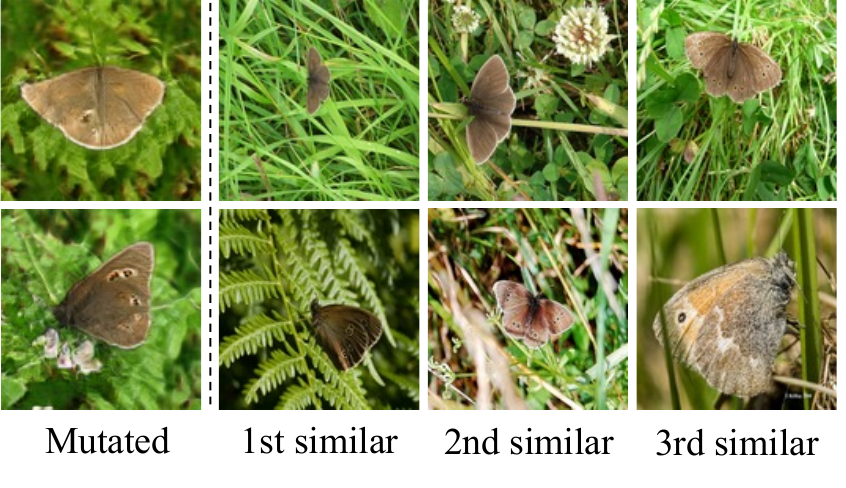}
  \vspace{-5pt}
  \caption{Mutated images and the most similar training data.}
  \label{fig:ssim-rq2}
  \vspace{-5pt}
\end{figure}

\noindent \textbf{\mr{Mutations, not Memorization.}}~DNNs are known for being able to memorize data
points. To show that \tool\ indeed performs perceptual-level mutations rather
than simply yields memorized training data, we calculate the similarity between the mutated
images in \F~\ref{fig:qualitative-rq1} and each training image using
SSIM~\cite{wang2004image}, a popular image similarity metric.
\F~\ref{fig:ssim-rq2} shows the top-3 most similar training images regarding two
mutated images. These training images still manifest distinct perceptions with
\tool's outputs, indicating that \tool\ constructs manifold to facilitate
perceptual-level mutations, rather than mere memorization.

\noindent \textbf{\mr{Mutated Perceptions.}}~\tool\ forms perceptual transformations
$\mathcal{T}$ when walking on manifolds.
Though manifolds have much lower dimensions (i.e., around 100), it is still hard
to interpret which perceptions are primarily mutated by $\mathcal{T}$ since
several perceptions are jointly mutated. We observe that in most cases,
perception mutations result in geometric-related changes, which is hardly
achievable by prior approaches; see more visualizations in~\cite{snapshot}.

\subsubsection{Diversity of Faults}

\mr{
Since different error-triggering inputs can be due to the same root cause,
we measure the diversity of these DNN faults to quantify and compare
the disclosed erroneous root causes: faults of higher diversity indicates more
erroneous root causes are revealed in testing.

\noindent \textbf{Erroneous Classes.}~Because all tested DNNs perform multi-label
classification, we first count how many classes are covered by the disclosed faults.
Results are given in \T~\ref{tab:error-rq2}. As noted, for ImageNet cases, \tool\
trigger DNN faults for most classes, which is around $3\times$ of the results
of TensorFuzz. For CIFAR10 cases, faults triggered by \tool\ cover all classes.
We interpret these results are highly promising: besides being
effective to disclose more DNN faults, these faults also manifest a higher
quality as they extensively reveal more diverse flaws.
}

\noindent \textbf{\mr{Scaled Entropy.}}~As reported in \T~\ref{tab:error-rq2}, since
CIFAR10 only has 10 classes, most previous tool have triggered faults for all
classes. Therefore, to further distinguish the diversity, we use scaled entropy as
a metric. The scaled entropy is $-\frac{1}{|C|}\sum p_{c}\log p_{c}$, where $|C|$ is
the number of classes and $p_{c}$
is the proportion of outputs incorrectly predicted to be class $c$. Scaled entropy
is in the range of $[0,1]$, and a higher entropy suggests more diversity, i.e., less
bias for particular classes. \T~\ref{tab:error-rq2} shows that \tool\ outperforms
others. We find that the scaled entropy is close to 1, indicating that findings of
\tool\ are unbiased and almost evenly distributed across all classes for all three
CIFAR10-DNNs.

\begin{table}[t]
  \caption{\mr{Diversity of disclosed DNN faults.}}
  \vspace{-5pt}
  \label{tab:error-rq2}
  \centering
\resizebox{0.88\linewidth}{!}{
  \begin{tabular}{
    @{\hspace{2.0pt}}c@{\hspace{2.0pt}}|
    @{\hspace{2.0pt}}c@{\hspace{2.0pt}}|
    @{\hspace{2.0pt}}c@{\hspace{2.0pt}}|
    @{\hspace{2.0pt}}c@{\hspace{2.0pt}}|
    @{\hspace{2.0pt}}c@{\hspace{2.0pt}}|
    @{\hspace{2.0pt}}c@{\hspace{2.0pt}}
    }
    \hline
     \multirow{2}{*}{}        & \multirow{2}{*}{Tool} & ImageNet   & CIFAR10    & CIFAR10   & CIFAR10      \\
                              &                       & ResNet-50  & ResNet-50  & VGG-16    & MobileNet-V2  \\                
     \hline
     \multirow{5}{*}{\#Class} &  \tool\               & \textbf{849} & \textbf{10}  & \textbf{10}  & \textbf{10}   \\
                              &  DeepHunter           & 691  & \textbf{10}  & \textbf{10}  & \textbf{10}   \\
                              &  TensorFuzz           & 297   & 9    & 5    & 5   \\
                              &  DeepSmart            & 524  & \textbf{10}   & 9   & \textbf{10}   \\
                              &  DeepTest             & 306   & \textbf{10}   & 9   & 7   \\
     \hline
     \multirow{5}{*}{\shortstack{Entropy\\($\times$100)}} &  \tool\         & N/A  & \textbf{99.05}  & \textbf{98.98}  & \textbf{98.84}   \\
                                                          &  DeepHunter     & N/A  & 73.91  & 69.13  & 69.04   \\
                                                          &  TensorFuzz     & N/A  & N/A    & N/A   & N/A  \\
                                                          &  DeepSmart      & N/A  & 94.76  & N/A   & 91.29   \\
                                                          &  DeepTest       & N/A  & 68.08  & N/A   & N/A   \\
    \hline
  \end{tabular}
  }
  \begin{tablenotes}
    \footnotesize
    \item \mr{*The diversity evaluation is conducted in a two-step approach:
    1) DNN faults are more diverse if the erroneous predictions cover more classes.
    2) When the number of covered classes are equal, a higher entropy indicates
    more diverse DNN faults.}
  \end{tablenotes}
 \vspace{-5pt}
\end{table}

\begin{table}[t]
  % \vspace{-10pt}
  \caption{\mr{Retraining accuracy (\%) of ResNet50.}}
  \vspace{-5pt}
  \label{tab:retrain-rq2}
  \centering
  \resizebox{1.0\linewidth}{!}{
  \begin{tabular}{
    @{\hspace{2pt}}c@{\hspace{2pt}}|
    @{\hspace{2pt}}c@{\hspace{2pt}}|
    @{\hspace{2pt}}c@{\hspace{2pt}}|
    @{\hspace{2pt}}c@{\hspace{2pt}}|
    @{\hspace{2pt}}c@{\hspace{2pt}}|
    @{\hspace{2pt}}c@{\hspace{2pt}}
    }
      \hline
      Init. & \tool\                   & DeepHunter      & TensorFuzz      & DeepSmart         & DeepTest  \\
      \hline
      CIFAR10 & \textbf{94.07\%}$\uparrow$(All) & 93.86\%$\uparrow$ & 93.83\%$\uparrow$ & 93.73\%$\downarrow$ & 93.24\%$\downarrow$ \\
      93.77\% & \textbf{93.96\%}$\uparrow$(2K)  & (All)             & (All)             & (All)               & (All)       \\ 
      \hline
      ImageNet & \textbf{83.28\%}$\uparrow$(All) & 78.86\%$\uparrow$ & 78.32\%$\uparrow$ & 72.90\%$\downarrow$ & 72.16\%$\downarrow$   \\
      76.15\%  & \textbf{80.75\%}$\uparrow$(7K)  & (All)             & (All)             & (All)               & (All)       \\
      \hline 
  \end{tabular}
  }
  \begin{tablenotes}
    \footnotesize
    \item \mr{*Results are averaged over 5 runs with maximum standard deviations
    $<10^{-4}$.} 
  \end{tablenotes}
  \vspace{-10pt}
  \end{table}

\subsubsection{Usefulness for Retraining}
\label{subsubsec:retrain}

To ``repair'' DNNs, it's common to retrain them using error-triggering inputs with ground
truth labels~\cite{wang2020metamorphic,zhang2021autotrainer}.
\mr{This process can be viewed as a way of performing data augmentation for the tested
DNN~\cite{shorten2019survey,yun2019cutmix}.}
\T~\ref{tab:retrain-rq2} shows the test accuracy after retraining of ResNet50 on
CIFAR10 and ImageNet. For each tool, we take all of their discovered error-triggering
inputs for retraining. Since \tool\ finds much more error-triggering inputs, to make the
comparison fair (DeepHunter has 2,011 and 7,455 error-triggering inputs for CIFAR10
and ImageNet), we randomly selected 2K and 7K error-triggering inputs of \tool\
for retraining on CIFAR10 and ImageNet, respectively.

For both ImageNet and CIFAR10, \tool\ surpasses others in both settings, i.e.,
83.28 (All) \& 80.75 (7K) and 94.07 (All) \& 93.96 (2K), indicating that the
flaws detected by \tool\ are of high quality and diversity. It empirically
reflects the \vall\ of perceptual-level mutations performed by \tool. We also
emphasize that real-world DNNs may have been trained using standard data
augmentation approaches like affine transformations~\cite{shorten2019survey}.
Therefore, faults triggered by \tool\ are more desirable and orthogonal to those
that may have been included in the augmented training data. It also explains why
TensorFuzz retraining had the similar accuracy as DeepHunter, even though
TensorFuzz triggered fewer faults.

\noindent \textbf{\mr{Prior Tools Harm the Retraining.}}~Retraining on datasets
augmented by DeepSmart and DeepTest results in a decrease in accuracy. The
region-wise mutation can break objects in images (see \F~\ref{fig:qualitative-rq1}),
and DeepTest may apply several affine modifications to an image, making it
less recognizable. As a result, the mutated images may no longer retain the
label and the follow-up retraining is therefore harmed.
This evaluation shows the merit of manifold-based mutation; in comparison with
other tools, \T~\ref{tab:retrain-rq2} empirically illustrates that \vall\
(perceptual constraints) is retained by \tool, given that retraining on
error-triggering inputs can notably enhance DNNs and outperform others.

\mr{
\noindent \textbf{Lower Improvements for Better DNNs.}~As a reported in
\S~\ref{subsec:effectiveness}, DNNs w.r.t. CIFAR10 are already well-trained to
a saturating performance (i.e., the test accuracy no longer increases when
trained with the original training data). Thus, an increased accuracy around
$0.5\%$ is indeed a big improvement --- it is $5\times$ of the improvement
made by the best competitor (i.e., DeepHunter). In contrast, for the ResNet50
trained on ImageNet, since performing a 1000-class classification is more
challenging, the initial accuracy is relatively lower. Accordingly, much
higher improvements are made by retraining with error-triggering inputs of \tool\
(i.e., +7.13\%); the gaps with improvements brought by other tools are also enlarged.
}

\subsection{Generalizability}
\label{subsec:generalizability}

\parh{Settings.}~This section benchmarks the generalizability of \tool.
\T~\ref{tab:dnn} lists the models and datasets. The manifold of driving
scenes has 512 dimensions and the text manifold has 100 dimensions. Since
both these two datasets only have one class, we construct one manifold
for each dataset. The audio dataset has 10 class; we construct 10 manifolds
and the dimension is 128. Each of the experiment in this section runs
for six hours.

\begin{table}[t]
  \caption{Quantitative results of generalizability.}
   \vspace{-10pt}
  \label{tab:quantitative-rq3}
  \centering
  \resizebox{0.80\linewidth}{!}{
    \begin{tabular}{
      @{\hspace{2pt}}c@{\hspace{2pt}}|
      @{\hspace{2pt}}c@{\hspace{2pt}}|
      @{\hspace{2pt}}c@{\hspace{2pt}}|
      @{\hspace{2pt}}c@{\hspace{2pt}}}
      \hline
      Task               & Tool      & Comprehensiveness & \#Faults (NC) \\                
      \hline
      \multirow{3}{*}{\shortstack{Autonomous\\driving}} &  $\text{DeepRoad}_\text{rain}$ &  (+6.15, +0.58, +0.82)  & 2375 \\
     &  $\text{DeepRoad}_\text{snow}$ &  (+7.36\%, +16.14\%, +2.32\%) & 2213 \\
     &  \tool\   &  (+\textbf{18.64\%}, +\textbf{25.94\%}, +\textbf{9.95\%}) & \textbf{15547} \\
     \hline
     \multirow{2}{*}{\shortstack{Machine\\translation}}   &  \multirow{2}{*}{\tool}   & \multirow{2}{*}{+9.96\%}                & \multirow{2}{*}{4011}  \\
        & & &  \\
     \hline
     \multirow{2}{*}{\shortstack{Audio\\classification}} &  \multirow{2}{*}{\tool}   & \multirow{2}{*}{+4.04\%}                & \multirow{2}{*}{1449}  \\
       & & &  \\
    \hline                
  \end{tabular}
  }
%  \vspace{-5pt}
\end{table}

\begin{figure}[!ht]
  \centering
  %\hspace{-15pt}
  \includegraphics[width=0.8\linewidth]{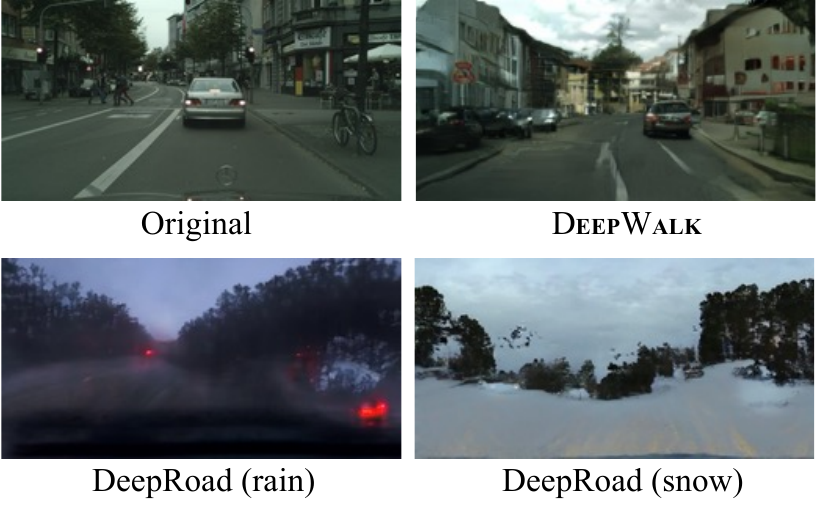}
  \vspace{-10pt}
  \caption{(Mutated) driving scene images.}
  \label{fig:qualitative-rq3}
  \vspace{-5pt}
\end{figure}

\subsubsection{Autonomous Driving --- Regression \& Differential Testing}
\label{subsubsec:driving}

\noindent \textbf{\mr{\mr{Differential Testing.}}}~Since the steering angle is likely
to change when perceptions of a driving scene are mutated by \tool, we set
up a differential testing scenario to form our oracle, in which steering angles
predicted by three DNNs must match. The testing objective for this setting is
the sum of values calculated on three target DNNs. The GAN model takes roughly
one day for training due to complexity of driving scenes. The tested DNNs are
provided by DeepTest~\cite{tian2018deeptest}
with pre-trained parameters.

\noindent \textbf{\mr{Comparison with DeepRoad.}}~We compared \tool\ with DeepRoad. The mutated outputs are presented in
\F~\ref{fig:qualitative-rq3}. In addition to adjusting the road direction,
\tool\ may also change the position/color of the car and background buildings,
which introduces diverse driving scenes.
We find that DeepRoad breaks original images when changing weathers. Therefore,
its metamorphic oracle on steering angle consistency does \textit{not} always
hold. We also assess DeepRoad in the differential testing setting. In
\T~\ref{tab:quantitative-rq3}, we use NC to reflect the testing comprehensiveness
and the number of faults are listed. DeepRoad uses additional data from other domains to deliver
``style translation.'' For instance, DeepRoad with a ``rainy'' scheme requires
1) seed driving scenes and 2) topologically similar scenes that offer rainy and
sunny scenarios. Although DeepRoad is enhanced with additional knowledge,
\tool\ outperforms it in both the testing comprehensiveness and \#faults.
\mr{Also, when NLC is used as the testing objective, 32,939 faults are triggered
by \tool, which is more than twice of the \#faults when using NC. This observation
is consistent with our results in \S~\ref{subsec:effectiveness} --- NLC is more
effective to guide generating error-triggering inputs. Note that unlike other
baseline tools, DeepRoad is not a feedback-driven tool; its triggered faults do
not change with testing objectives.}

\mr{ It's worth mentioning that, because the tested task is regression whose
outputs are continuous, measuring the diversity of faults using entropy or
\#classes is infeasible. Nevertheless, since \tool\ comprehensively explores
different perceptual-level mutations (e.g., mutating the background building or
the car's color; see more examples in~\cite{snapshot}) and DNNs generally rely
on perceptual properties in inputs to make predictions, it should be accurate
to assume that faults triggered by \tool\ are due to more and different root
causes. In addition, while analyzing root causes for DNN faults is still an open
problem, we envision that perceptual-level mutations enabled by \tool\ can benefit future
research for the interpretation and root cause analysis of DNN faults
(see more discussions in \S~\ref{sec:future}).

\noindent \textbf{Connection to Physical-World Attack/Testing.}~Previous works
have conducted attacks/testings for autonomous driving systems via physical-world
mutations~\cite{zhou2020deepbillboard,sato2021dirty} (e.g., adding printable
adversarial billboard~\cite{zhou2020deepbillboard}), which are more likely
to occur in real world. Since manifolds in \tool\ are approximated in real-life
images, we view the perceptual-level mutation in \tool\ as a potential direction
for extending physical-world mutations. By approximating manifolds over real-life
images, developers can observe which perceptual properties the autonomous driving
system is sensitive to (i.e., the prediction can easily change when mutating
these properties). These sensitive properties denote fault triggers that broadly
exist in physical world and have more severe threats. With \tool, developers can
repair the autonomous driving system in a more oriented manner. 
}

\subsubsection{Audio --- More Complex Media Data}
\label{subsubsec:audio}

\parh{Settings.}~GAN for preparing manifold is built
from stacked CNNs (see~\cite{snapshot}) whose training completes in 2 hours.
Audio clips of the same number being spoken lie in one manifold. The target
DNN was pre-trained with $93.96\%$ test accuracy on the SC09 dataset. NC is
used to reflect the testing comprehensiveness and results are given in
\T~\ref{tab:quantitative-rq3}.

\parh{\mr{Faults \& Retraining.}}~We note that a considerable number of
faults are triggered. Overall, mutating audios is changing given the complex
format. \tool\ directly mutates perceptions in audios which does not rely
on any specific format and is very effective to trigger DNN faults.
We further retrain the tested model on SC09 augmented with error-triggering
inputs. The test accuracy improves to $94.52\%$, demonstrating the high quality
of the mutated (error-triggering) audio recordings. The improvement is not high,
which is reasonable because the tested DNN is already well-trained.
Overall, we interpret the improvement in retraining as proving that \tool\
captures perceptual constraints on audios and produces realistic audios.
We present mutated audios from all manifolds in~\cite{snapshot}.

\subsubsection{Machine Translation --- New Task \& Discrete Data}

\noindent \textbf{Settings.}~\tool\ mutates natural language texts
in a grammatically coherent manner, without prior knowledge or predefined
templates~\cite{galhotra2017fairness,udeshi2018automated,ma2020metamorphic}. The
procedure is the same as mutating image or audio, except that we first
use ARAE to encode discrete text into a continuous representation (discussed in
\S~\ref{subsec:gan}). The whole training pipeline takes less than 2 hours. We use
NC as the criteria.
We test two jointly trained MT models, namely $\vec{E2G}$ and $\vec{G2E}$, that
translate English (German) to German (English). Since MT primarily focuses on
grammatical norms of languages, all English text can be viewed as being in one
manifold.

\begin{figure}[!ht]
  \centering
  %\vspace{-10pt}
  \includegraphics[width=0.75\linewidth]{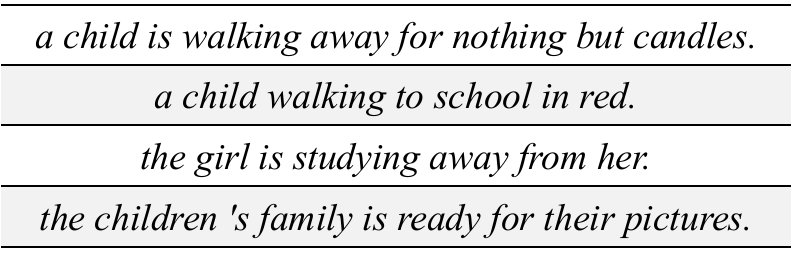}
 %\vspace{-10pt}
  \caption{Mutated sentences from the same seed.}
% \vspace{-10pt}
  \label{fig:text}
\end{figure}

\parh{\mr{Grammatical Coherence.}}~\tool\ could build grammatically coherent yet different English sentences
by walking on the constructed manifold.
Empirically, we adopt the fluency score~\cite{ge2018fluency,ma2020mt} to
assess coherence of mutated sentences. The fluency score, higher the better, ranges
from $[0, 1]$ and measures how fluent a sentence is via word dependencies.
All mutated sentences achieve a fluency score of 0.32, suggesting strong
coherence, compared to 0.35 for normal phrases in SNLI. We present several
sample sentences in \F~\ref{fig:text}, which are mutated from the same seed.
They are grammatically diverse and coherent, despite the fact that \tool\ does
not specify or require templates to mutate texts. More examples of mutated
sentences are shown in~\cite{snapshot}.

\parh{\mr{Back Translation.}}~It's generally hard to decide if MT makes translation errors. We form the
testing oracle by assessing the consistency of $\vec{E2G}$ and $\vec{G2E}$. We
convert an English sentence $eng$ to a German sentence $ger$ using $\vec{E2G}$
and then back to an English sentence $eng'$ using $\vec{G2E}$; $eng'$ should be
highly similar with $eng$. An inconsistency indicates that at least one of the
models translate incorrectly.

\parh{\mr{Testing Oracle \& Results.}}~We assess the similarity of $eng$ and $eng'$ via BLEU (Bi-Lingual Evaluation
Understudy)~\cite{papineni2002bleu}. The BLEU score is highly correlated with
human assessment and ranges from $[0, 1]$ (higher is better). We denote $eng$ as
an error-triggering input if the BLEU score between $eng$ and $eng'$ is $0$.
The sentence
``\underline{A black dog carrying a metal to a man}'' is a fault-triggering
sentence produced by \tool. This sentence is translated as ``\underline{Ein schwarzer Hund
h$\ddot{\text{a}}$lt einen Mann an}'' and is back-translated to ``\underline{A black dog is
holding a man}''. \T~\ref{tab:quantitative-rq3} reports the testing comprehensiveness
and \#faults. \tool\ triggers a great number of faults in six hours.
The results are encouraging, showing \tool's adaptability in handling diverse
media data types.

\begin{table}[t]
  %\vspace{-12pt}
  \caption{Results of quantized DNN.}
  \vspace{-5pt}
  \label{tab:result-rq4}
  \centering
\resizebox{0.65\linewidth}{!}{
  \begin{tabular}{l|c|c}
    \hline
    Quantized layer & Test acc. (\%) & \#Fault \\
    \hline
     Conv2D              & 58.37    & 139,122  \\ 
    \hline
     Linear              & 58.32    & 146,793 \\
    \hline
     Linear + Conv2D     & 58.32    & 165,065 \\ 
    \hline
  \end{tabular}
  }
\end{table}

\subsubsection{Quantized DNNs --- A Black-Box Scenario}

We also simulate a testing scenario for
DNNs on mobile devices, where the DNN is quantized. We employ a MobileNet-V2
pretrained on ImageNet (officially shipped by PyTorch) and the quantization
schemes are listed in \T~\ref{tab:result-rq4}. All three schemes quantize the
layer parameters from 32-bit floating number to 8-bit integer. We present the
test accuracy (on the randomly select $100$ classes from ImageNet) in
\T~\ref{tab:result-rq4}. The original model has $58.59\%$ test accuracy,
and the accuracy drops induced by quantization are negligible. The quantized
models are thus deemed equivalent to the unquantized one. We use \tool\ to
test these quantized models.

\parh{\mr{Black-Box Entropy.}}~In this setting, applying white-/gray-box objectives is infeasible, since we
only have the DNN output, i.e., a predicted label and the associate probability.
We thus adopt the black-box entropy as the testing objective, which maximizes
the difference of predicted probabilities from the DNN and the quantized one.
Accordingly, we set up a differential testing oracle and denote a fault as an
inconsistent prediction between the original and quantized DNN.
The \#faults are listed in \T~\ref{tab:result-rq4}. \tool\ produces a vast number of
inconsistency-triggering inputs which reveal hidden defects introduced by quantization.

\mr{
\section{Threat to Validity}
\label{sec:threat}

\parh{Bias in Evaluations.}~Since this research compares \tool\ with
prior tools, one threat is that the evaluation is biased. To mitigate
this threat, we select a considerable number of DNNs that are widely
studied in previous works~\cite{pei2017deepxplore,tian2018deeptest,xie2018coverage,yuan2022unveiling,yuan2023revisiting},
and commonly adopted as the backbone of modern deep learning systems.
These 19 DNNs are representative in terms of tasks, structure, working scenarios,
and featured platforms. We also consider five popular real-world datasets
which consist of various data formats including image, audio, and text.
In particular, the ImageNet dataset has been deemed as the ``golden standard''
of benchmarking DNNs for the past ten years. DNN testing objectives in our
evaluation are also representative (e.g., structural, cluster-based, and
distribution-aware) and proposed by research from both the AI and
SE communities.

\parh{False Positives.}~Given that we report the faults disclosed by \tool,
another threat is that there might be false positives, for instance, invalid
inputs or mutated inputs whose ground truth labels are changed. We clarify that
this issue should be eliminated from the following aspects.
1) Conceptually, data manifold aims to capture perceptual-level constraints of
media data and \tool\ is based on manifolds provided by existing mature techniques.
2) Technically, manifolds are separately constructed for data of different
labels and test inputs that are likely violate perceptual-level constraints
(i.e., having high sensitivity; see \S~\ref{sec:formal}) are pruned before testing.
3) Empirically, as evaluated in \S~\ref{subsubsec:eval-validation}, \tool\ nevers
generates format-invalid test inputs whereas prior tools frequently violates the
format restriction of input. Also, \S~\ref{subsubsec:recognizability} demonstrates
that test inputs generated using \tool\ manifest
the highest recognizability. Moreover, in \S~\ref{subsubsec:retrain}, we retrain
the tested DNNs: when using error-triggering inputs of \tool, the retrained DNNs
have the highest improvements. Thus, it is reasonable to infer that
false positives are largely eliminated in \tool's results.

\parh{False Negatives.}~The third threat faced by \tool\ is that some DNN
faults can be neglected. We clarify that, like all prior works in this
field, \tool\ aims to test DNN rather than verify DNNs: it cannot guarantee
to disclose all DNN faults. Nevertheless, we deem \tool\ as a powerful
testing tool as it enables effective perceptual-level mutations and simultaneously
retains the validity of mutated inputs. \tool\ manifests a high generalizability:
it can test DNN of various 1) input formats (e.g., image and audio),
2) tasks (classification and regression), and 3) deployed
scenarios/platforms (gray-box and black-box), under 4) different testing
oracles (metamorphic and differential testing).

\section{Limitations and Future Works}
\label{sec:future}

\parh{Fine-Grained Annotations.}~Perceptual-level mutations may break fine-grained
annotations on DNN inputs. For instance, as shown in \S~\ref{subsec:generalizability},
when testing autonomous driving, the steering angle will be altered if the road in a
driving scene is mutated via perceptual-level mutations. This threat can potentially
harm the generalizability of perceptual-level mutations for (metamorphic) testing of
DNNs under certain scenarios.
Nevertheless, this issue is not specifically induced by \tool; geometrical mutations
(since steering angle is tied to geometrical properties) and the knowledge transfer
in DeepRoad (see examples in \F~\ref{fig:qualitative-rq3}) can also change the ground
truth steering angle in the mutated images.

On the other hand, for previous tools performing mutation on data bytes
(e.g., pixel-level mutations), although they may produce invalid and unnatural data,
they can reuse the dense annotation in most cases if the operator and validator are
carefully designed by human experts.

Recall as we show in \F~\ref{fig:manifold}, previous mutations explore
the $\epsilon$-radius sphere around the original input. Thus, to alleviate the above threat,
it is possible to retain the human annotations by restricting the footsteps when \tool\ is 
walking on manifold---with the diversity of mutation is unavoidably sacrificed.
In addition, users may construct ``sub-manifolds'' within one manifold, for instance,
given a collection of driving scenes, users can construct a ``sub-manifold'' that
corresponds to driving scenes of directing-to-left road. We leave these as
future works.

\parh{Testing Objective.}~Most testing objectives aim to capture the
behaviors/states of DNNs~\cite{pei2017deepxplore,ma2018deepgauge,yuan2023revisiting}.
Nonetheless, understanding internal mechanisms of DNNs is still an open problem,
which impedes further developments of DNN coverage. Since DNNs make prediction
based on perceptual properties in inputs, we view proposing perception coverage
metrics as a promising research direction. That is, instead of counting states
of DNNs, we can count how many perceptual properties are covered during testing.
Unlike existing coverage metrics which may depend on how tested DNNs are
implemented, perception coverage is agnostic to the implementations of DNNs,
because it is defined in the input space. In addition, perception coverage is
orthogonal to existing coverage metrics that focus on DNN behaviors. In that
sense, it can be combined with existing coverage metrics to form a hybrid
coverage to better characterize the testing comprehensiveness.

\parh{Testing Oracle.}~Existing works mainly define testing oracles for DNN
outputs, i.e., whether the DNN (or multiple DNNs) have consistent predictions
before/after mutations. One recent work advocates to define oracle over the
decision process of DNNs~\cite{yuan2022unveiling} (i.e., decision oracle).
Despite viewing from different angles, they do not distinguish DNN faults.
Therefore, we deem forming perceptual-level oracle as an interesting future
direction. More concretely, the oracle checks whether the DNN makes consistent
prediction before/after mutating certain perceptual properties. Compared with
existing oracles, perceptual-level oracles are more fine-grained, as it separates
DNN faults according to the mutated perceptual properties. Moreover, perceptual-level oracles
are orthogonal to existing output- and decision-based oracles; perceptual-level oracles can also be
combined with prior oracles to form hybrid ones and 
reflect DNN faults in a faith manner.

\parh{Interpretable Root Cause Analysis.}~We view employing \tool\ for interpretable
root cause analysis of DNNs as a highly feasible research direction. Compared
with previous mutations (e.g., adding noise), perceptual-level mutations enabled
by \tool\ are more understandable and interpretable for human developers. For
instance, when a DNN fault is triggered after mutating the ear in a dog photo,
the developer can conclude that the DNN focuses on ears to recognize dogs and
accordingly fix this bias. Moreover, since \tool\ maintains $\mathcal{T}$ to
record its footsteps on data manifolds, by visualizing mutations depicted by
$\mathcal{T}$, developers can holistically decide which perceptual properties
the DNN is most vulnerable to.
}

\section{Conclusion}
\label{sec:conclusion}

This research formulates the key objectives in mutating 
DNN inputs, \divv\ and \vall. We prove that \divv\
and \vall\ inextricably bounds each other, and rebut SOTA works on their
applicability of mutating real-world media data. We then present \tool\ for mutating
media data to achieve high \divv\ and \vall\ with provably guarantee.
Our evaluation shows \tool's high effectiveness, exceeding prior works in both general image
classification and domain-specific scenarios.

\bibliographystyle{IEEEtran}
\bibliography{bib/main}

\end{document}